\def\year{2021}\relax
\documentclass[letterpaper]{article}
\usepackage{aaai22}
\usepackage{times}
\usepackage{helvet}
\usepackage{courier}

\usepackage[utf8]{inputenc}

\usepackage{natbib}
\usepackage{caption}

\usepackage[hyphens]{url}
\usepackage{graphicx}
\graphicspath{ {figures/} }

\usepackage{tabularx}
\usepackage{amsmath}
\usepackage{subcaption}
\usepackage{booktabs}
\usepackage{multirow}

\title{Are You Robert or RoBERTa? Deceiving Online Authorship Attribution Models Using Neural Text Generators}
\author{Keenan Jones, Jason R.~C.~Nurse, Shujun Li\\}
\affiliations{
Institute of Cyber Security for Society (iCSS) \& School of Computing\\ 
University of Kent, UK \\
\{ksj5, j.r.c.nurse, s.j.li\}@kent.ac.uk}

\begin{document} 

\maketitle 

\begin{abstract} 

Recently, there has been a rise in the development of powerful pre-trained natural language models, including GPT-2, Grover, and XLM. These models have shown state-of-the-art capabilities towards a variety of different NLP tasks, including question answering, content summarisation, and text generation. Alongside this, there have been many studies focused on online authorship attribution (AA). That is, the use of models to identify the authors of online texts. Given the power of natural language models in generating convincing texts, this paper examines the degree to which these language models can generate texts capable of deceiving online AA models. Experimenting with both blog and Twitter data, we utilise GPT-2 language models to generate texts using the existing posts of online users. We then examine whether these AI-based text generators are capable of mimicking authorial style to such a degree that they can deceive typical AA models. From this, we find that current AI-based text generators are able to successfully mimic authorship, showing capabilities towards this on both datasets. Our findings, in turn, highlight the current capacity of powerful natural language models to generate original online posts capable of mimicking authorial style sufficiently to deceive popular AA methods; a key finding given the proposed role of AA in real world applications such as spam-detection and forensic investigation.

\end{abstract}

\section{Introduction}

With the proliferation of powerful natural language models like GPT-2/3, Grover, and XLM~\cite{uchendu2020}, we have seen rapid advancements made in the field of natural language generation (NLG). Using large pre-trained models fine-tuned with minimal amounts of training data, these natural language models have shown state-of-the-art capabilities in generating coherent and stylised texts~\cite{radford2019}. Moreover, the current capacity of these state-of-the-art models is such that it is becoming ever more difficult for humans to distinguish between human-created and AI-generated texts~\cite{uchendu2020}. Parallel to this, there has been an significant amount of work dedicated to the subject of online authorship analysis. That is, the development of models (typically utilising machine learning) to successfully identify the author of a given online post~\cite{kaurOSM2019}. 

However, less work has been conducted examining the interaction between AI-based text generation and authorship analysis. In this research we present, to the best of our knowledge, the first efforts to study the abilities of powerful natural language models to automatically generate online texts that mimic a target authorial style such that they can deceive online authorship attribution (AA) models. To this, we answer the following research questions (RQs):

\begin{itemize}
\item \textbf{RQ1}: To what degree are texts automatically generated by neural natural language models able to deceive state-of-the-art AA models?

\item \textbf{RQ2}: How does the originality of the AI-generated text affect the degree to which AI-generated texts can successfully deceive state-of-the-art AA models?

\item \textbf{RQ3}: To what extent do AI-generated texts capture the intrinsic stylistic qualities of human-created texts from the author they attempt to mimic?
\end{itemize}

Through this analysis, applied to both blog and Twitter data, we note that pre-trained natural language models are capable of automatically generating texts that capture the intrinsic stylistic patterns of a given author. Additionally, we find that these AI-generated texts are successful in deceiving state-of-the-art AA models, which frequently attribute AI-generated texts to the given author they attempt to mimic. Moreover, we also find that this capability of AI-generated AA deception holds true even when the amount of fine-tuning data used to generate the deceptive texts is minimal.

These findings highlight the current capabilities of powerful natural language models towards deceptive text generation, and hint towards the potential capabilities (and dangers) of greater authorial mimicry in the near future as these models grow in capability.

\section{Related Work}

AA, the form of authorship analysis we focus on in this paper, involves the matching of an anonymous document to its most likely author from a pre-established set of candidate authors~\cite{kaurOSM2019}. This approach is typically treated as a multi-class classification problem, where each author in the set is treated as a discrete class~\cite{kaurOSM2019}. 

With the proliferation of machine learning, deep learning, and powerful pre-trained natural language models, the potential for authorship analysis is higher now than it ever has been, and the ability for researchers to extract more information from short-form online text sets has improved substantially. This has led to studies being carried out in the use of AA on blog posts~\cite{fabien2020},  tweets~\cite{daneshvar2018GenderII}, Facebook status updates~\cite{ding2017, hu2020}, and dark market listings~\cite{kaurOSM2019}. More sophisticated approaches have also been proposed beyond the conventional methods of using classical machine learning classifiers trained on either stylistic or n-gram features, leveraging neural networks and pre-trained natural language models to learn more sophisticated embedding representations of authorial style~\cite{fabien2020, hu2020}. Further study has also focused on the use of AA in cross-domain scenarios, successfully leveraging models trained on the posts of an author on one online platform to identify posts by the same author on a different platform. This has included work on controlled corpora of blog, emails and interviews~\cite{barlas2020} and fanfiction~\cite{kestemont2020}.

Parallel to this, recent years have seen the rise of powerful natural language models such as BERT, GPT-2/3, and Grover~\cite{uchendu2020}. Pre-trained on vast amounts of data, these models -- typically built around a transformer architecture~\cite{vaswani2017} -- have shown state-of-the-art capabilities in a variety of natural language processing (NLP) tasks~\cite{radford2019}.

Despite the popularity of these powerful natural language models, and the capacity of many of them for high quality text generation, little research has been conducted to examine the implications of this for authorship analysis. As authorship analysis research is often framed in terms of its practical applications, including digital forensic investigations~\cite{perkins2018}, detecting comprised online accounts~\cite{kaurOSM2019}, and detecting phishing emails~\cite{duman2016}, it is of crucial importance that the implications and potential weaknesses of authorship analysis methods when faced with powerful natural language models are considered.

In \cite{uchendu2020}, the authors examined the ability of AA methods to determine whether a given news article is written by a human, or produced by one of a series of powerful text generators. They found that typical authorship methods were able to achieve good performances when trained specifically for the task of distinguishing between human-created and AI-generated texts and the task of identifying the responsible model of a given AI-generated text~\cite{uchendu2020}.

In a related field, recent work has been conducted in the field of author-style transfer using pre-trained natural language models~\cite{jin2020}. In this task, the aim is to obscure or transfer the authorial style of a given input text whilst retaining its semantic content. In \cite{Syed2020}, the authors utilised powerful natural language models to rewrite a given text without the need for parallel data, leveraging a cascade of natural language models in an encoder-decoder framework for authorial style transfer. The authors also proposed novel metrics for measuring the degree of authorial stylistic alignment, leveraging lexical and syntactical patterns to score similarities in writing styles. Moreover, in \cite{goyal2021}, the authors combined multiple `style aware' natural language models to allow for the transferal of both sentiment and formality. This approach was able to outperform that of \citet{Syed2020} in terms of style transference, though not in terms of content preservation, and achieving strong performs in both of these aspects remains a challenge.

In addition to these papers, other research has focused on deceiving authorship analysis methods using a variety of non-generative approaches. In \cite{mahmood2020}, the authors examined the task of authorship obfuscation -- the attempt to disguise texts by a given author so that they remain undetected by authorship analysis methods. In turn, the authors experimented with a variety of state-of-the-art authorship obfuscation methods, including the use of genetic algorithms to identify words in a given text that would have the highest effect in obscuring the source author when substituted~\cite{mahmood2019}. Document simplification approaches have also been considered, which utilise rule-based text simplifications, including replacement of words with synonyms, and contraction and expansion replacement, to disguise authorship~\cite{castro2017}. Style neutralisation techniques have also been examined in this context, leveraging the stylometric properties of a text (e.g,. the average number of words per sentence, punctuation to word count ratio) to move a given input text's stylometric features closer to that of the average features of the corpus as whole, thereby `smoothing' the text and removing an author's stylistic inputs~\cite{karadzhov2017}.

\section{Our Contributions}

Our work presents the first attempts to examine the capacity of powerful natural language models to automatically generate new online texts capable of capturing a given user's writing style in order to deceive AA methods.

The current work most related to ours, author-style transfer~\cite{Grondahl2020,Syed2020}, focuses on rewriting existing inputs -- preserving a given input text's latent semantic content whilst transferring or obscuring its authorial style. In our work, however, we focus instead on the degree to which natural language generators used to automatically generate new texts (as opposed to rewriting existing texts) can imitate the style of a given author.

Additionally, whilst these recent works have identified the capacity of powerful natural language models to transfer stylistic attributes~\cite{Syed2020,goyal2021}, they have generally leveraged bespoke metrics to measure the extent to which transference was achieved. Thus, the degree to which natural language model-based style transference is capable of deceiving AA models also remains unstudied. Given the relatively large amount of focus that AA has received in recent years, this of particular interest.

Beyond these studies, the work most similar to ours lies primarily in authorship obfuscation, which is a similar but fundamentally different task to the work conducted here. In obfuscation, the aim is to change a given text to hide its author from a given AA system; in our work we focus on authorship imitation, where we attempt to create new texts that mimic a target author in order to fool AA classifiers.

Whilst it is true that existing work has been conducted to develop models capable of detecting AI-generated texts, it is worth noting that these solutions are typically limited in their generalisability. Generally, successful models are only effective in detecting AI-generated texts from a specific \emph{known} model (e.g., GPT-2, XLM) of a specific type (e.g., news articles, blog posts)~\cite{jawahar2020}. There exists no ``silver bullet'' capable of making these potential deceptive texts trivial to identify. Moreover, as authorship analysis models are often framed as tools that can offer real-world practical uses~\cite{kaurOSM2019}, including in forensics, spam and phishing detection, and the identification of compromised accounts, an understanding of any potential vulnerabilities in these approaches is therefore crucial. Whilst methods exist that may be useful in combating the threat of authorial generative deception, an awareness of these threats is still needed in order to determine what safeguards may be required when developing AA models. 

Our work highlights this potential vulnerability, and advocates for greater consideration of the weaknesses of current AA approaches in detecting AI-generated deceptive texts to ensure that they are appropriately protected. While the inconsistent nature of current text generators weakens the capacity of author-styled text generation, the rise of more powerful models with steerable outputs could increase these capabilities in the near future~\cite{dathathri2019}. It is therefore crucial that any potential weaknesses in these systems are identified early so mitigation strategies can be developed.

\section{Methodology}
\label{label:Method}

In this paper, we examine the ability of powerful natural language models to automatically generate online posts that retain the authorial style of a given user. We also conduct linguistic analyses to examine the degree to which text generators mimic authorial stylistic patterns. Our complete approach is detailed below.

\subsection{Data Collection}

We first needed to obtain relevant datasets containing a large number of posts from a range of authors. In order to obtain a good understanding of the ability of natural language models in this task we opted to focus on two different styles of online posts: blog posts and tweets. 

For the blog data, we used the popular Blog Authorship Corpus~\cite{schler2006}. Constructed from 681,288 English posts from 19,320 users on blogger.com, this corpus has seen frequent use in NLP challenges, including previous authorship analysis studies~\cite{fabien2020}.

For the Twitter dataset, we opted to curate our own corpus. To do this, we collected a random sample of Twitter users using Twitter's real-time sampling API~\cite{TwitterAPI}. This collection was conducted in March 2021 and resulted in a set of 11,698 Twitter accounts being sampled. Twitter's timeline API was then used to extract the latest tweets from each account. We then filtered any non-English tweets to ensure that language was standardised amongst users and removed all retweets. This yielded an initial dataset of 1,969,214 tweets from 10,924 accounts.

\subsection{Online Post Generation}
\label{label:Method;PostGen}

\begin{table*}[t]
\footnotesize
\centering
\begin{tabularx}{\linewidth}{ cccccc }
 \toprule
\textbf{Dataset} & \textbf{Total fine-tuning documents} & \textbf{Total posts generated} & \textbf{Avg.~posts per author} & \textbf{Avg.~post length} & \textbf{Avg.~token length}\\
\midrule
\multirow{5}{*}{Blog} 
& 50 & 5813 & 58.13 & 93.24 & 3.61\\
& 100 & 6527 & 65.93 & 97.29 & 3.61\\
& 150 & 7267 & 73.4 & 109.27 & 3.58\\
& 200 & 7122 & 71.94 & 103.28 & 3.6\\
& 250 & 7600 & 77.55 & 102.87 & 3.57\\
\midrule
\multirow{5}{*}{Twitter} 
& 50 & 8494 & 84.94 & 14.2 & 3.94\\
& 100 & 8623 & 86.23 & 14.23 & 3.94\\
& 150 & 8889 & 88.89 & 14.55 & 3.9\\
& 200 & 8845 & 88.45 & 14.31 & 3.89\\
& 250 & 9036 & 90.36 & 14.26 & 3.91\\
 \bottomrule
\end{tabularx}
\caption{Details of the generated datasets produced using GPT-2 and fine-tuning documents sampled from the 100 authors from each of our datasets.}
\label{table:generated_datasets}
\end{table*}

Having established our datasets of human-created texts, we then moved on to producing AI-generated texts. In order to test the ability of natural language models in capturing a given authorship style, this process therefore involved the fine-tuning of a series of natural language models on collections of online posts obtained from a single user. Having fine-tuned a given model using these posts, this model could then be used to generate new posts that, ostensibly, capture the authorial style of the intended user.

It was first necessary to sample from our two datasets the precise number of authors and posts per author that would be used. As some authors' posts may allow for better or worse generated outputs, it was necessary to include a number of authors in our experiments. We thus generated texts from 100 randomly sampled authors from each dataset. This number of authors would ensure that the results recorded were likely an indication of the performance of our text generation models on each platform, rather than the result of the model's performance on a very small set of authors. 300 posts were then randomly sampled per author from each dataset.

We then needed to consider the number of posts per author that would be used to fine-tune each model. As the number of posts used here would likely impact the quality of the AI-generated texts, we opted to test a range of different post numbers to examine the effect this would have. In turn, we decided to test values between 50 and 250 posts per author, with increments of 50 being tested (i.e., 50, 100, 150, 200, 250). These posts were obtained through randomly sampling from the total 300 posts per author.

To generate our texts, we used the GPT-2~\cite{radford2019} language model. Developed by OpenAI, GPT-2 is a popular natural language model that has seen frequent use in the literature. The model utilises a large transformer-based architecture with `1.5 billion parameters, trained on a dataset of 8 million web pages' allowing it to achieve strong performances in a range of NLP tasks~\cite{radford2019}.

For our experiments, we leveraged the large GPT-2 model with 774 million parameters~\cite{radford2019} using the gpt-2-simple (\url{https://github.com/minimaxir/gpt-2-simple}) Python package and fine-tuned each generative model using the next token prediction task~\cite{uchendu2020}. As our datasets contain relatively short texts, we followed standard conventions of delimiting each individual text leveraged during fine-tuning using ``$<$|startoftext|$>$'' and ``$<$|endoftext|$>$'' tokens. 

For the generation process, we largely relied on the default fine-tuning and generation hyperparameters. One exception to this is the number of training steps used during the fine-tuning stage. As the forms of data we used -- blog posts and tweets -- are relatively short in nature, we reduced the number of training steps from the default of 1,000 to avoid overfitting the model to the fine-tuning data. This is important as overfitting can lead to higher chances of duplication in content between the generated texts and the original set of fine-tuning texts. For both our blog and Twitter datasets, we experimented with different numbers of training steps to find the optimum values that avoid overfitting whilst still providing coherent generations. Through manual analysis, we identified that 500 steps had yielded the best results in this context. We also adapted the ``text length'' hyperparameter (used to guide the length of each generated text) to ensure each generated text's lengths best reflect the typical length of its respective author's writing. To identify the appropriate text length for each author, we use the average character length of the posts used during fine-tuning. As prompts for generation, we simply use the ``$<$|startoftext|$>$'' token that had been used to delimit each of the fine-tuning texts. By using a generic prompt, we attempt to avoid potential confounders of specific prompts being more or less suited to individual authors.

For each fine-tuned GPT-2 model we generated 100 texts. As the output of these text generators can often be inconsistent, this sufficiently large number helps ensure that a reasonable number of good-quality texts are produced. After generation, we conducted some initial filtering to remove low-quality texts. To achieve this, we removed any generated texts that either duplicated existing texts in their respective author's genuine writings, or that were exact copies of other generated texts. This ensured that all generated texts were unique. After this, we then removed any generated texts with fewer than 5 words. This was done as we reasoned that texts of such short length were less likely to be able to encode any sense of authorial style. Details regarding our generated datasets can be found in Table~\ref{table:generated_datasets}.

\subsection{Authorship Attribution (RQ1 \& RQ2)}

Having generated our texts, we then moved onto the core task of our research: AA. An AA task is centred around the assignation of a given input text to its `true' author from a discrete set of authors~\cite{kaurOSM2019}: $K$. Following previous research, we treat this as a multi-class classification problem~\cite{uchendu2020}, where a given machine learning model is trained and validated on texts drawn from the authors in $K$.

In our case, the AA task is assessed not in terms of the ability of a given model to assign genuine texts to the correct author, but instead its tendency to be fooled by AI-generated texts that mimic the author's style. To this, we train an AA model using genuine human-created texts and then test it on AI-generated texts (as discussed in Section~\ref{label:Method;PostGen}) to see if the generators can successfully deceive the AA model. If an AI-generated text intended to mimic target author \textit{A} is predicted to be from \textit{A}, than the attribution model has been successfully deceived. For all of our experiments, we used a fixed $K$ size of 5, as the previous literature indicated that this number allows state-of-the-art AA classifiers to have a good chance of achieving reasonable-to-good performances (accuracies $>$ 0.7) when trained on human-created data~\cite{fabien2020}. As the number of authors increases, the performances of AA classifiers will generally decrease drastically. This, in turn, would make it difficult to assess whether a classifier's `incorrect' classification of a deceptive generated post was due to the post not adequately capturing the target author's style, or as a result of the classifier inadequately learning to identify the author's writings in the first place.

Our first step was to pre-process both the human-created and AI-generated texts. As AA relies on the encoding of subtle patterns in each author's style, we opt for a minimalistic approach -- removing each author's name from their posts, replacing all URLs with a $<$\emph{URL}$>$ tag, and replacing any user mentions in our tweet data with a $<$\emph{USER}$>$ tag.

For our AI-generated texts, we additionally had to consider the possibility for generated texts to be duplicates or to share a high level of similarity with texts existing in the fine-tuning dataset. 
To account for this, we used Levenshtein distance. Levenshtein distance measures the similarity of two sequences by examining the number of edits required to change one sequence into the other~\cite{su2008}. We tested a series of different Levenshtein distance thresholds, removing any generated posts with a Levenshtein distance above the given threshold when compared against each of the parent author's fine-tuning posts. We experiment with the affect that using different Levenshtein thresholds to filter our generated texts have on our attribution models' performance.

Our next step was to select the models and features that would be used for AA. We began by experimenting with classical machine learning models combined with stylometric and $n$-gram based features. For our stylometric features, we generated a large feature set of 791 stylometric features, drawn from the existing literature~\cite{kaurOSM2019}. Examples of these feature types can be found in Table~\ref{table:features}. Where relevant, we recorded both raw counts and normalised measurements. Moreover, we also utilised both word and character $n$-gram features using values of $n$ from 1 -- 3. Pointwise mutual information (PMI) was then used to select a subset of the most useful features. For our experiments we selected three of the most popular AA classifiers~\cite{kaurOSM2019}: random forest (RF), decision trees, and support vector machines (SVM) with a linear kernel. Python Scikit-learn implementations were used for each of the machine learning classifiers tested~\cite{Pedregosa2011}.

\begin{table}[h!]
\footnotesize
\centering
\begin{tabular}{l} 
 \toprule
\textit{Character Features} \\
- Number of uppercase characters \\
- Avg. word length \\
\textit{Word Features} \\
- Number of uppercase words \\
- Number of dictionary words \\
\textit{Sentence Features} \\
- Avg. words per sentence \\
- Number of sentence beginning with uppercase characters \\
\textit{Lexical Diversity Features} \\
- Yule's K \\
- Simpson's D \\
 \bottomrule
\end{tabular}
\caption{Examples of stylometric features extracted from our datasets.}
\label{table:features}
\end{table}

Alongside this, we also experimented with the use of AA classifiers based on language models. Whilst the use of pre-trained natural language models for AA has not received a significant amount of study, recent studies have indicated that they are capable of state-of-the-art performances~\cite{fabien2020}. Given its noted performance in previous AA tasks~\cite{fabien2020}, we experimented with the popular BERT model~\cite{devlin2018}, leveraging both the base model and -- for our Twitter dataset -- the RoBERTa-based BERTweet model~\cite{nguyen2020}. The Hugging Face Python package was used for all natural language model-based attribution (\url{https://huggingface.co/}).

\subsection{Linguistic Analysis (RQ3)}

We also conducted a series of linguistic comparative analyses of the AI-generated texts relative to their human-author-created counterparts. Firstly, we utilised the popular Linguistic Inquiry and Word Count (LIWC) tool~\cite{tausczik2010}. LIWC utilises a series of dictionaries containing words related to a specific lexical or psychological `dimension'~\cite{tausczik2010}. By analysing the proportion of words belonging to a given dictionary, LIWC provides a measure of the presence of a given dimension in a text. We compared the LIWC dimensions scores of human-created posts from a given author to that of AI-generated posts mimicking the same author.

To do this, for each author in our blog and Twitter datasets we concatenated together all of their human posts into a single document and all of the AI-generated posts produced from the same author's posts into a single document. We then used LIWC to record scores for each human and generated document. Thus, for each author in each dataset we produced two vectors of LIWC scores. As these LIWC vectors encode scores for a wide range of dimensions, we then distilled each LIWC vector into sub-vectors based on their dimensional categories~\cite{tausczik2010}. These sub-vectors allowed for more meaningful comparisons between the LIWC scores of human-created and AI-generated texts of the same author. Cosine similarity values were then used to compare the scores from a given LIWC category for human-created and AI-generated posts of the same author. 

Alongside these language analyses we conducted a comparative analysis of the topics present in AI-generated and human-created posts of the same author to examine the degree to which the AI-generated texts captured the topics present in the author's texts.

To do this, we applied the popular latent Dirichlet allocation (LDA) topic modelling approach. LDA works under the assumption that each given document is comprised of a random mixture of latent topics, with each topic being made up of a particular distribution of words~\cite{Blei2003}. A trained LDA model is able to provide both a probability distribution of words over topics in a corpus of documents and a probability distribution of topics over a specific document.

\begin{figure*}[tb]
\centering
\begin{subfigure}[b]{0.42\textwidth}
    \centering
    \includegraphics[width=\textwidth]{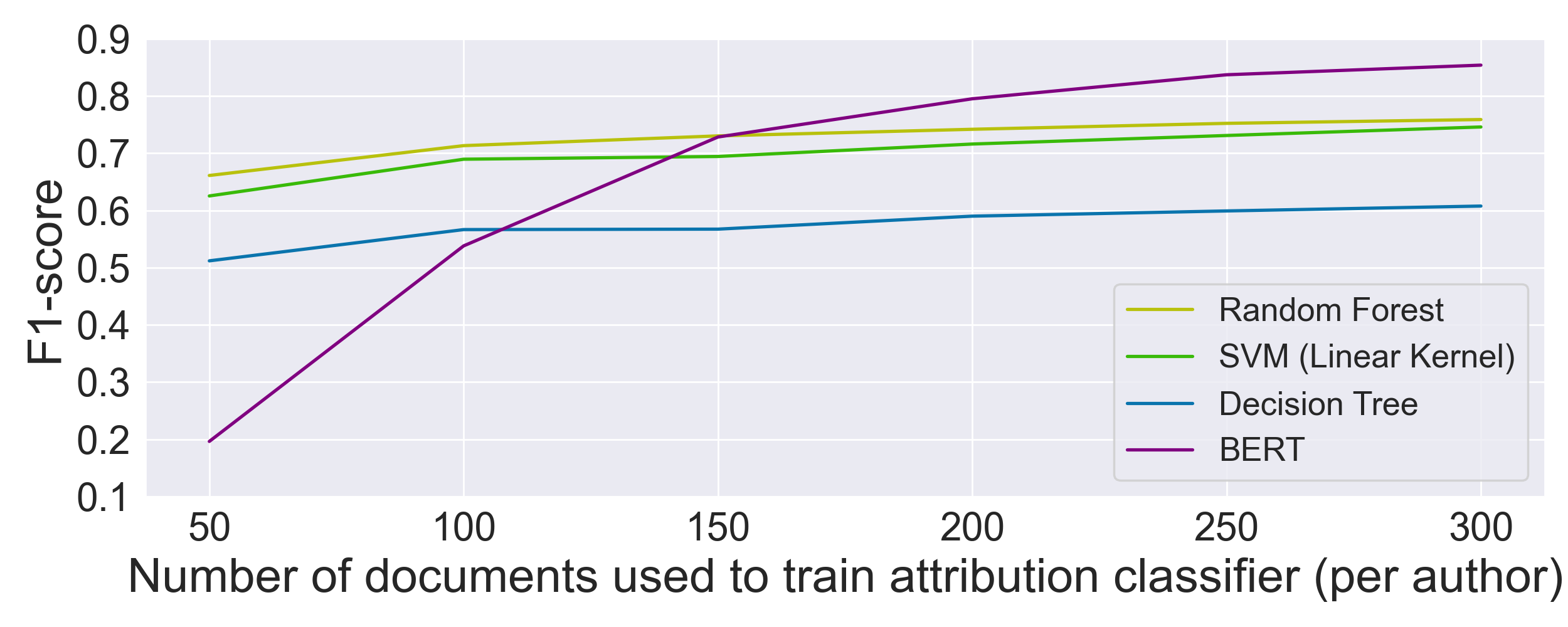}
    \caption{Results for the blog dataset.}
    \label{fig:human_cross_val_blog}
\end{subfigure}
\quad
\begin{subfigure}[b]{0.42\textwidth}
     \centering
     \includegraphics[width=\textwidth]{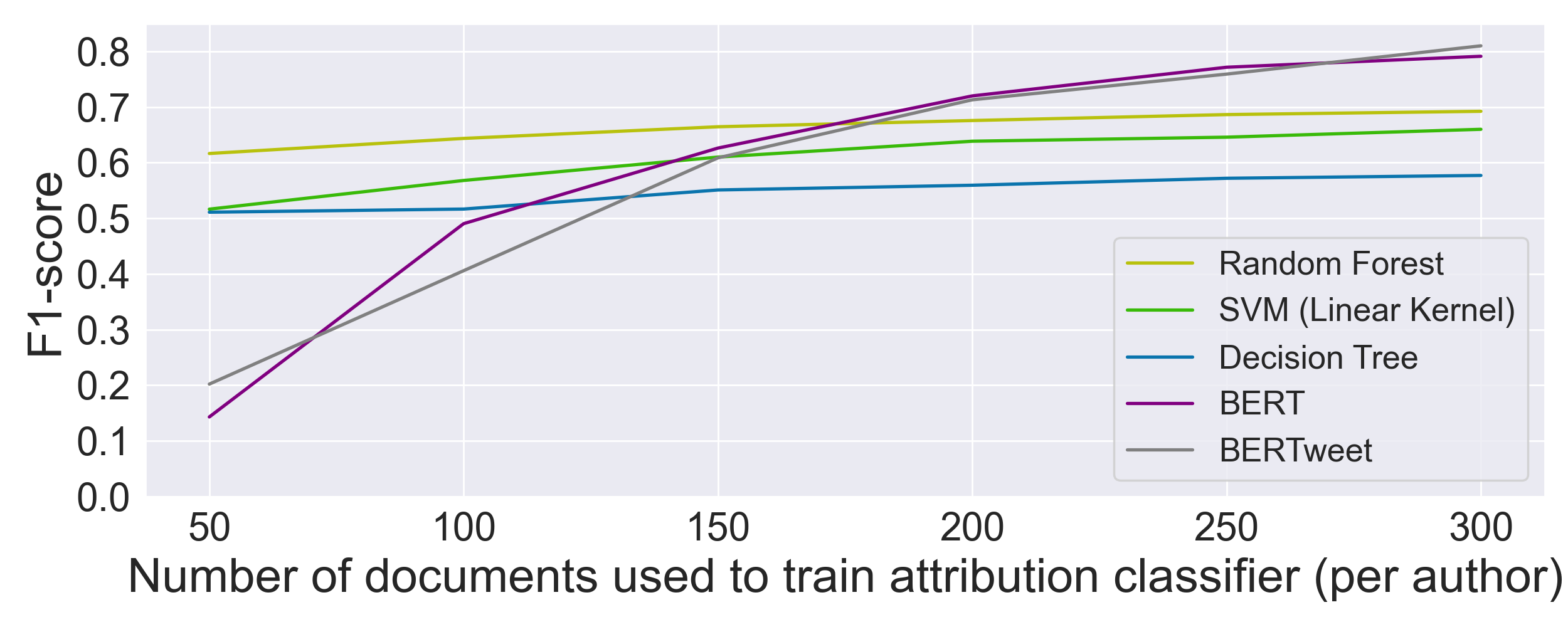}
     \caption{Results for the Twitter dataset.}
     \label{fig:human_cross_val_twitter}
\end{subfigure}
\caption{Authorship attribution F1-scores when cross-validated on our datasets of genuine posts, for different sizes of the training set.}
\label{fig:human_cross_val}
\end{figure*}

Firstly, we divided the authors in each of our two corpora into 20 trials, where each trial contained 5 randomly sampled authors (5 was chosen, as it is the standard size of $K$ used for our AA analysis). For each trial, we created a sub-corpus of the human-created and AI-generated posts for each of the 5 authors. Each sub-corpus was then used to build an LDA model, allowing it to learn the latent topics present across the sub-corpus of genuine and generated posts. As LDA topic modelling requires the number of topics $k$ that the model will distribute over, we utilised an iterative tuning method to identify the optimum $k$ for each trial. To do this, we used the popular UCI topic coherence metric, which provides a measure of the quality of the topics learned at a given value of $k$~\cite{Roder2015}. For each trial we iterated through a set of values for $k$ from 5 to 55 topics in steps of 5, constructing for each $k$ value an LDA topic model and recording its UCI score. We then identified at each trial the topic number that produced the optimum UCI score, which was then used to create our final set of 20 LDA models (per dataset).

Our LDA models were then used to generate the topic probability distributions for each human-created and AI-generated post for each author. We then examined similarities between the topic post probability distributions from AI-generated and human-created texts for the same author. For each author in each trial, we took the mean of the probability distributions of their human-created posts and of their AI-generated ones. This yielded two summative topic probability vectors for each author, one for their AI-generated posts and one for their human-created posts. Normalised Euclidean similarity values (calculated as $1 - \text{Normalised Euclidean Distance}$) were then utilised to measure the degree of similarity in the paired topic probability vectors for each author.

\section{Results and Discussion}
\label{label:Results}

\subsection{Authorship Attribution on Human-Created Posts}
\label{label:Results;AAHuman}

Before examining the degree to which natural language model-based text generators are capable of mimicking authorship, it is essential that we achieve a baseline understanding of how our AA models perform on each dataset.

In Table~\ref{table:human_AA_performances}, we present the precision, recall, and F1-scores for the four different AA classifiers on both datasets (blogger.com and Twitter). All results were obtained via the use of 10-fold cross-validation.

\begin{table}[!htb]
\small
\centering
\begin{tabular}{ c c c c c }
\toprule
\textbf{Dataset} & \textbf{Model} & \textbf{Precision} & \textbf{Recall} & \textbf{F1-Score} \\
\midrule
\multirow{4}{*}{Blog} & BERT (base) & \textbf{0.87} & \textbf{0.86} & \textbf{0.86}\\
& Random forest & 0.77 & 0.76 & 0.76 \\ 
& SVM & 0.76 & 0.75 & 0.75\\
& Decision tree & 0.62 & 0.61 & 0.61\\
\midrule
\multirow{4}{*}{Twitter} & BERTweet & \textbf{0.82} & \textbf{0.8} & \textbf{0.8}\\
& BERT (base) & 0.79 & 0.78 & 0.78 \\ 
& Random forest & 0.72 & 0.72 & 0.71 \\ 
& SVM & 0.68 & 0.68 & 0.67\\
& Decision tree & 0.62 & 0.62 & 0.62\\ 
\bottomrule
\end{tabular}
\caption{Authorship attribution cross-validation performances on human data for blog and Twitter datasets.}
\label{table:human_AA_performances}
\end{table}

\begin{figure*}[!tb]
\centering
\begin{subfigure}[b]{0.44\textwidth}
    \centering
    \includegraphics[width=\textwidth]{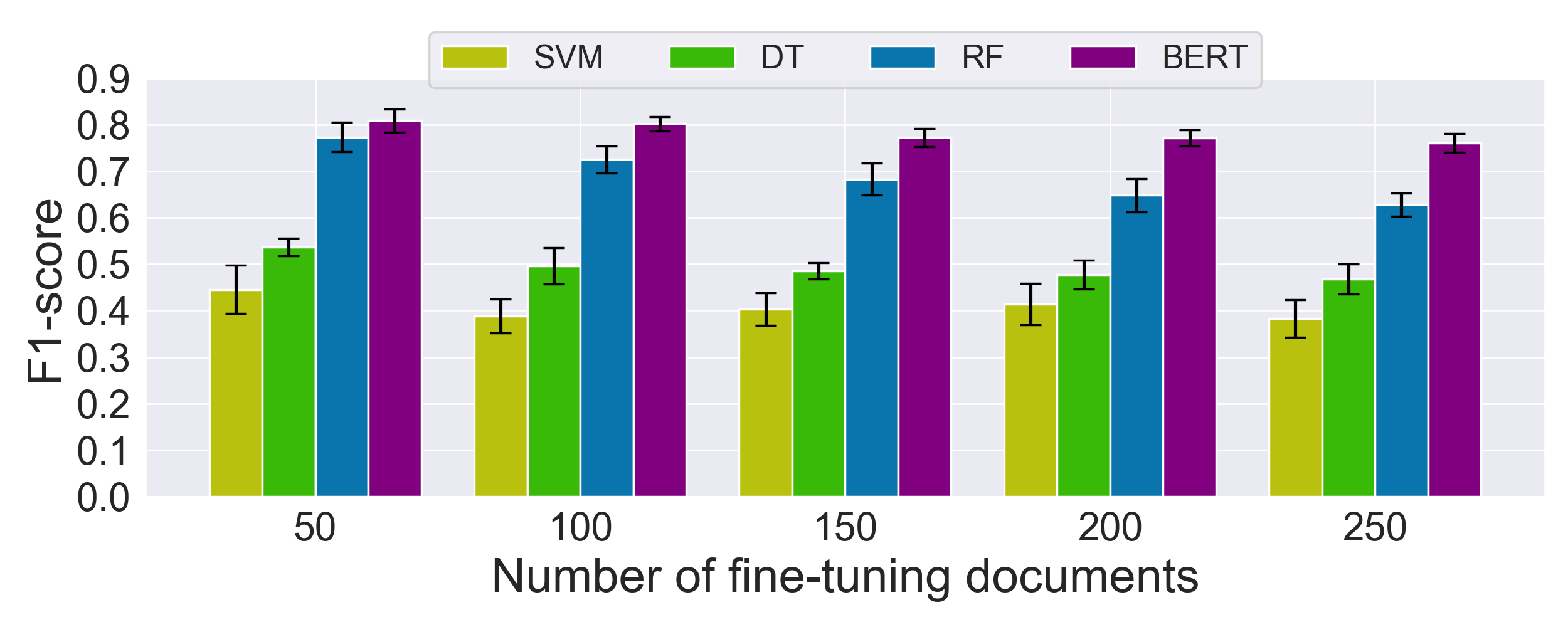}
    \caption{Results for the blog dataset.}
    \label{fig:human_train_gen_test_blog}
\end{subfigure}
\quad
\begin{subfigure}[b]{0.44\textwidth}
     \centering
     \includegraphics[width=\textwidth]{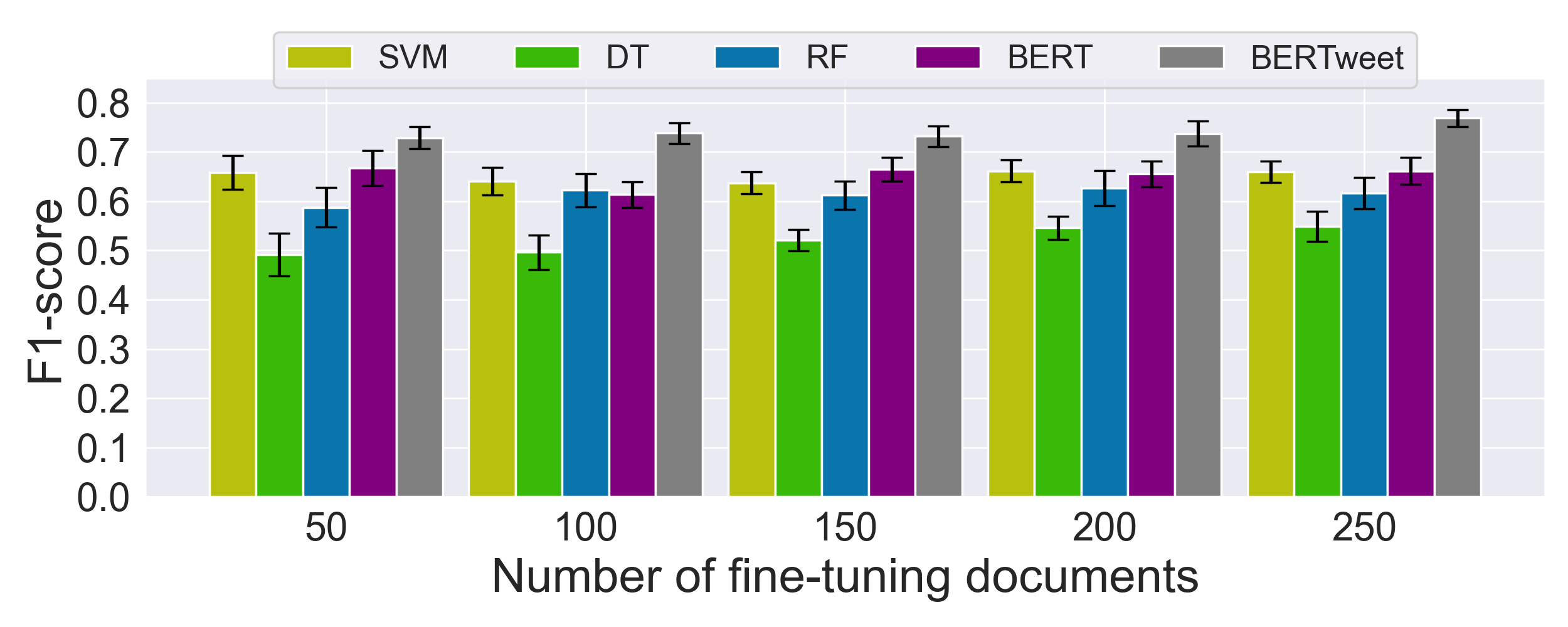}
     \caption{Results for the Twitter dataset.}
     \label{fig:human_train_gen_test_Twitter}
\end{subfigure}
\caption{Authorship attribution performances for models trained on genuine texts and tested on generated texts for increasing numbers of GPT-2 fine-tuning documents (per author).}
\label{fig:human_train_gen_test}
\end{figure*}

From this, we find that of the three classical machine learning classifiers, RF and SVM achieved similarly good performances on the Blogger.com (hereafter, the blog dataset) and similarly fair performances on the Twitter dataset. We note that in contrast, decision trees appear to perform distinctly worse on both datasets.

Moreover, the BERT models outperformed all three classical models on both datasets. Additionally, we note that the BERTweet model achieved higher performances than those of the BERT base model on the Twitter dataset. However, this increase in performance relative to the base BERT model does appear to be incremental compared to the differences in performance between natural language models and classical machine learning classifiers.

We also examined how varying the number of training documents affected AA model performance. These results can be found in Fig.~\ref{fig:human_cross_val}. In the interest of space we only report F1-score, but we can confirm that precision and recall scores were highly similar to the F1-scores in all cases.

In terms of the classical machine learning models, we find that the initial results in Table~\ref{table:human_AA_performances} are continued, with RF and SVM performing similarly well and decision trees performing markedly worse. We also note that for all of our classical models performance appears relatively stable as the number of documents are varied, increasing by around 10\% on both datasets as the training documents are increased.

Interestingly, this stability is not present for our BERT models. Instead, the natural language models performs very poorly with smaller numbers of documents per author, only exceeding RF's performances around 150--200 documents per author.

\subsection{RQ1: Using AI-based Text Generation to Mimic Authorship}

\begin{figure*}[tb]
\centering
\begin{subfigure}[b]{0.42\textwidth}
    \centering
    \includegraphics[width=\textwidth]{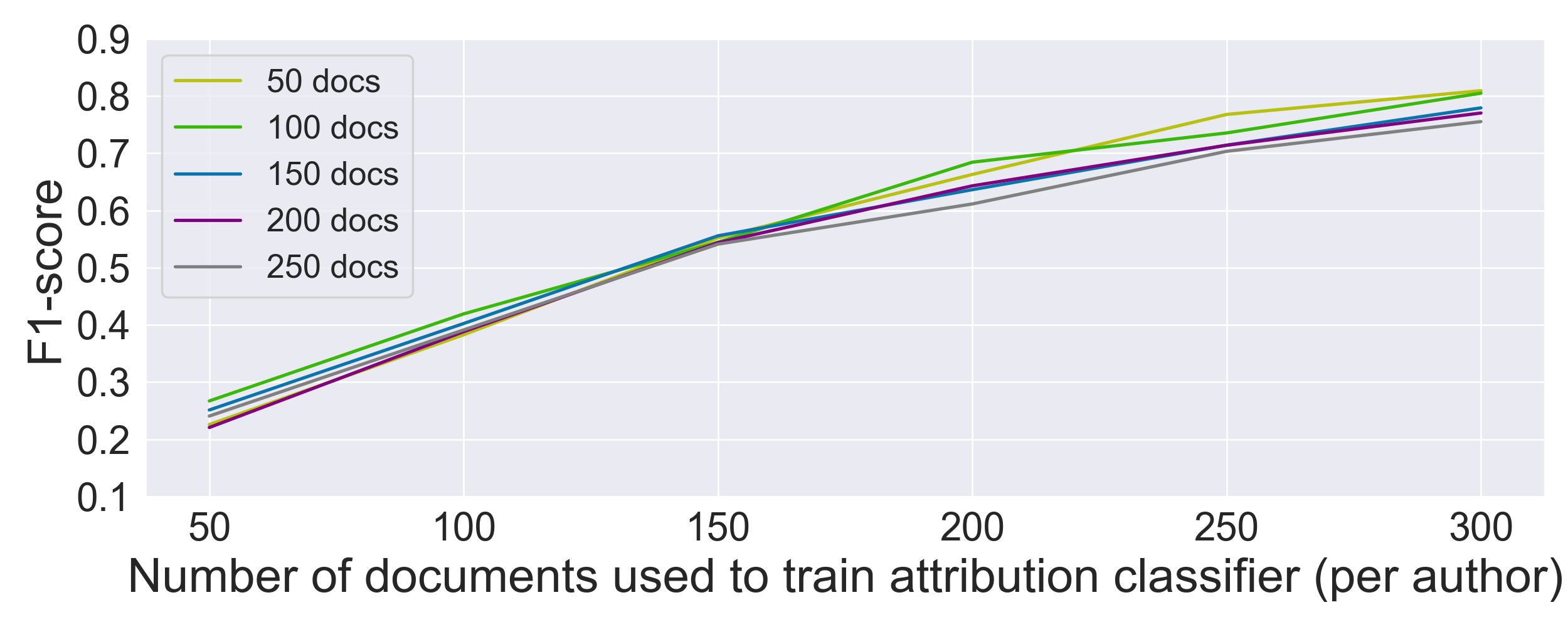}
    \caption{Results on the blog dataset using the BERT attribution model.}
    \label{fig:gen_test_blog_bert_varying_docs}
\end{subfigure}
\quad
\begin{subfigure}[b]{0.42\textwidth}
     \centering
     \includegraphics[width=\textwidth]{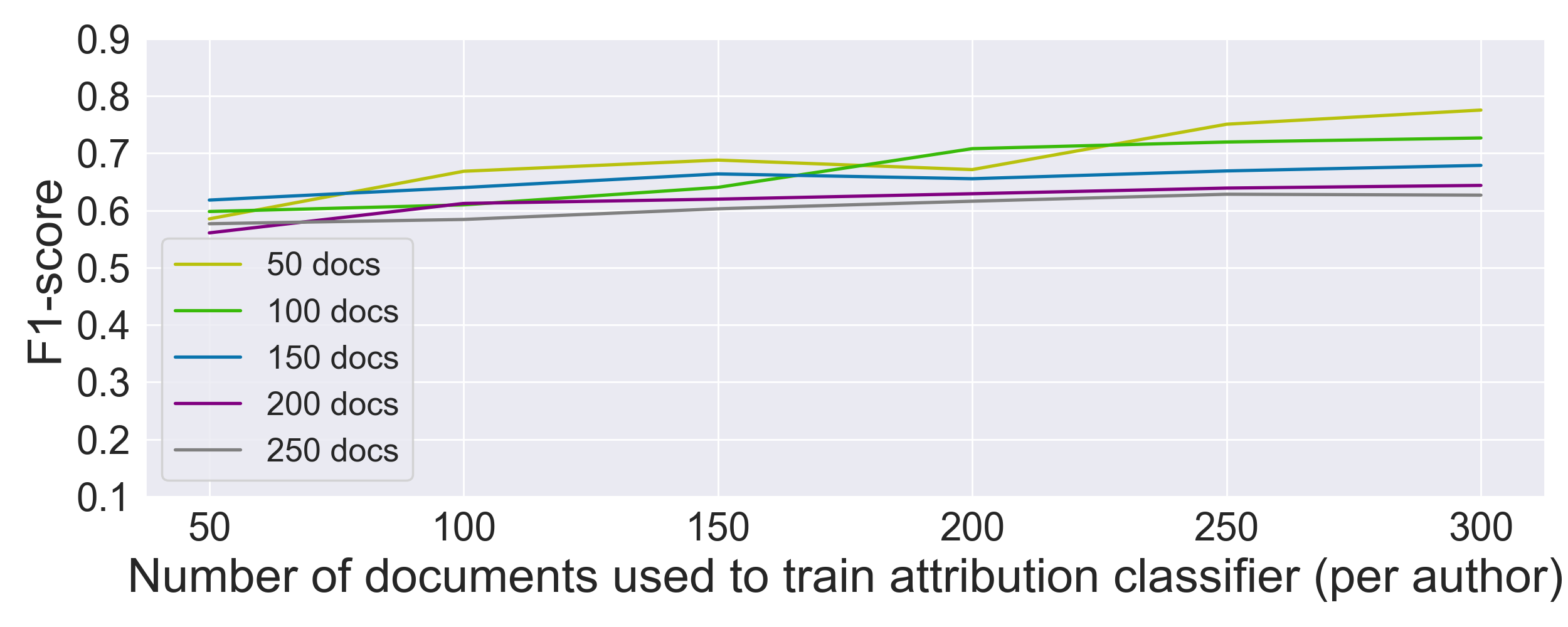}
     \caption{Results on the blog dataset using the random forest attribution model.}
     \label{fig:gen_test_blog_rf_varying_docs}
\end{subfigure}
\begin{subfigure}[b]{0.42\textwidth}
     \centering
     \includegraphics[width=\textwidth]{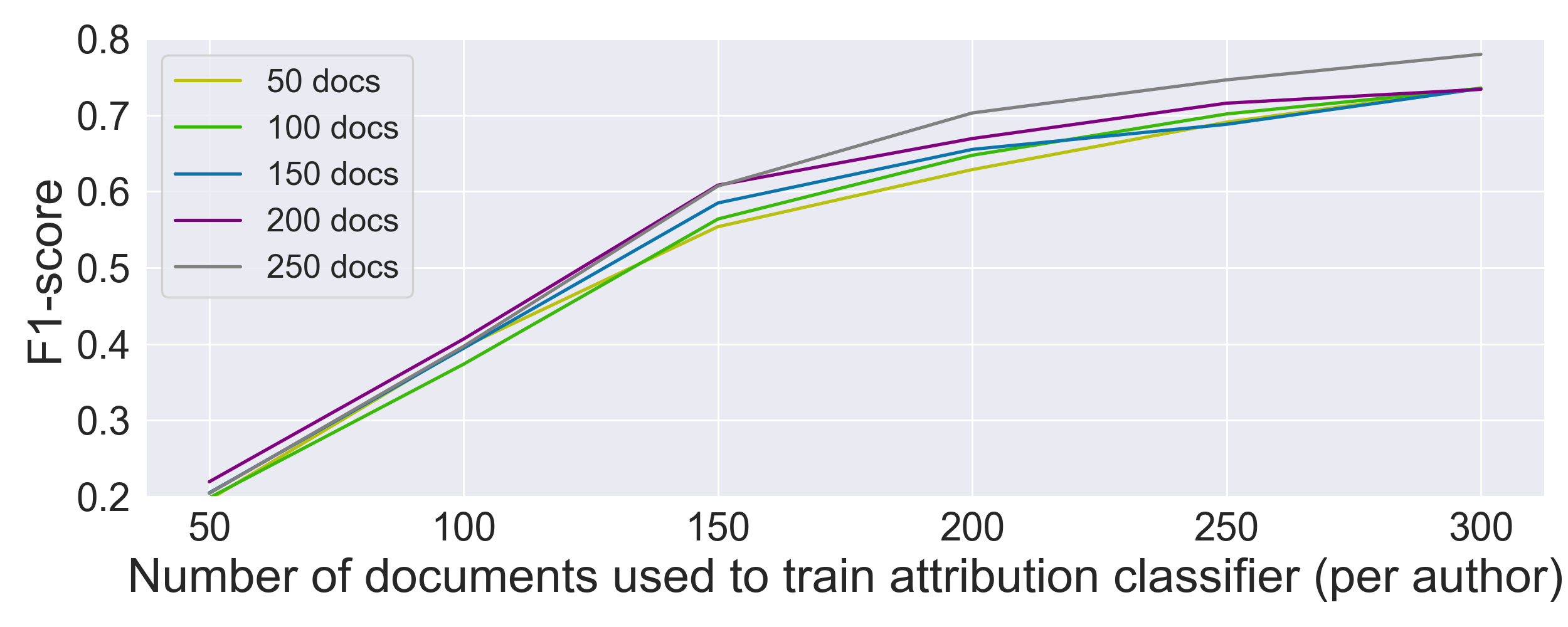}
     \caption{Results on the Twitter dataset using the BERTweet attribution model.}
     \label{fig:gen_test_twitter_BERT_varying_docs}
\end{subfigure}
\begin{subfigure}[b]{0.42\textwidth}
     \centering
     \includegraphics[width=\textwidth]{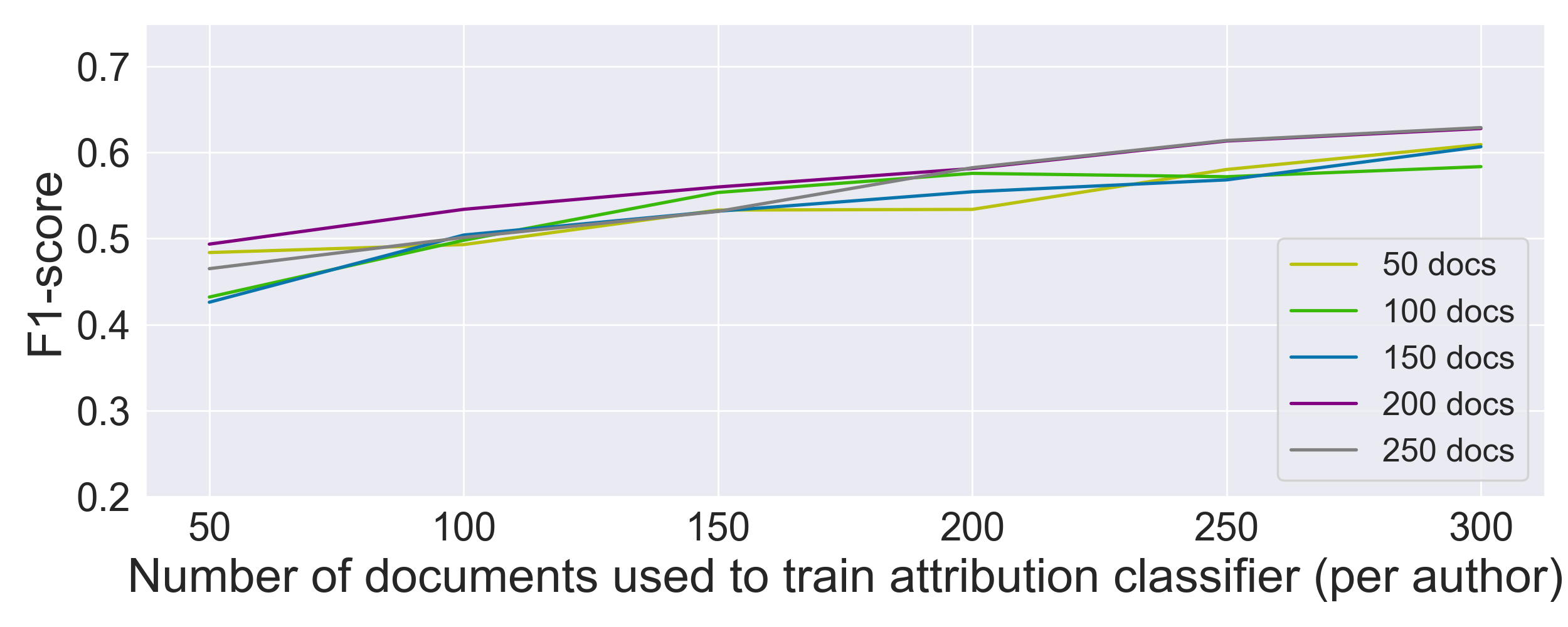}
     \caption{Results on the Twitter dataset using the random forest attribution model.}
     \label{fig:gen_test_twitter_rf_varying_docs}
\end{subfigure}
\caption{Authorship attribution performances for models trained on human-created texts and tested on AI-generated texts for increasing numbers of training documents for the attribution model. Legend indicates the different number of fine-tuning documents used per author.}
\label{fig:gen_test_varying_docs}
\end{figure*}

We now proceed to the core question underlying our research: can AI-generated texts, produced by models fine-tuned on human-created texts, deceive AA models?

For our initial experiments, we replicated the approach taken for the models in Table~\ref{table:human_AA_performances}, using 20 trials, sets of 5 authors, and 300 human-created training documents per author. Having trained each AA model, we then assess its performance on the AI-generated texts produced via fine-tuning using the documents from each of the 5 authors.

The results of these initial experiments can be found in Fig.~\ref{fig:human_train_gen_test}. In the interest of space, we do not present recall or precision scores, though we can confirm that in all cases these were highly similar to the F1-score.

From this, we find that on the blog data both RF and BERT, the models with the highest performances in Section~\ref{label:Results;AAHuman}, consistently attribute AI-generated texts to the `correct' authors with F1-scores upwards of 0.8. It thus seems that the AI-generated blog texts encode authorship in such a way as to be convincing to the BERT and RF models.

We also note that varying the number of fine-tuning documents seems to have minimal effect on the models' performances, particularly in regard to the BERT classifier. Instead, it appears that even with only 50 fine-tuning documents GPT-2 is still capable of generating posts that mimic authorship sufficiently for deception.

Interestingly, whilst SVM achieved similar performances to our RF model on the human-created datasets, it achieves very low scores on AI-generated posts for the blog dataset. This is interesting, as SVM on the human-created blog data achieved strong scores. Despite this, SVM does not show the susceptibility to AI-generated texts that the BERT and RF based models do, achieving consistently low scores of $\sim$0.4. It thus seems that whilst SVM is not be capable of the performances of RF or BERT, it may not be so easily `convinced' by the AI-generated texts.

\begin{figure*}[!htp]
\centering
\begin{subfigure}[b]{0.42\textwidth}
    \centering
    \includegraphics[width=\textwidth]{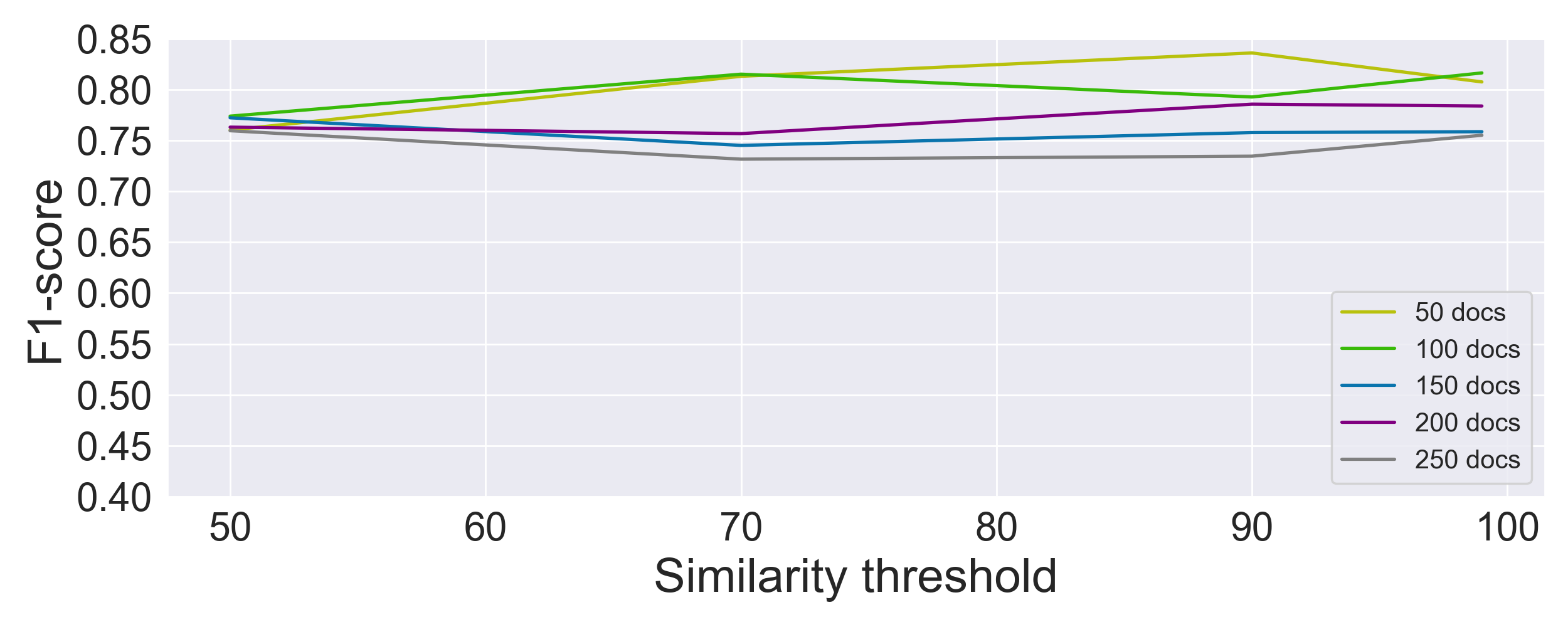}
    \caption{Results on the blog dataset using the BERT attribution model.}
    \label{fig:gen_test_blog_bert_varying_sim}
\end{subfigure}
\quad
\begin{subfigure}[b]{0.42\textwidth}
     \centering
     \includegraphics[width=\textwidth]{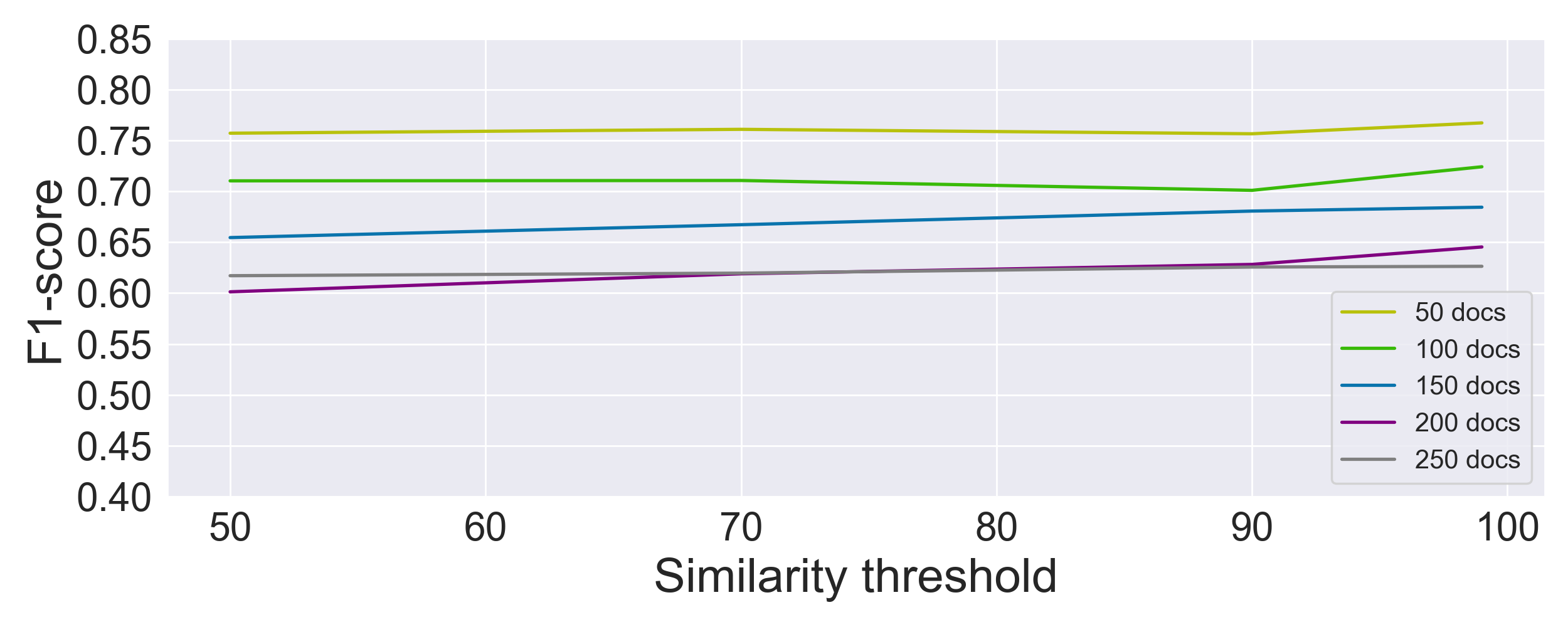}
     \caption{Results on the blog dataset using the random forest attribution model.}
     \label{fig:gen_test_blog_rf_varying_sim}
\end{subfigure}
\begin{subfigure}[b]{0.42\textwidth}
     \centering
     \includegraphics[width=\textwidth]{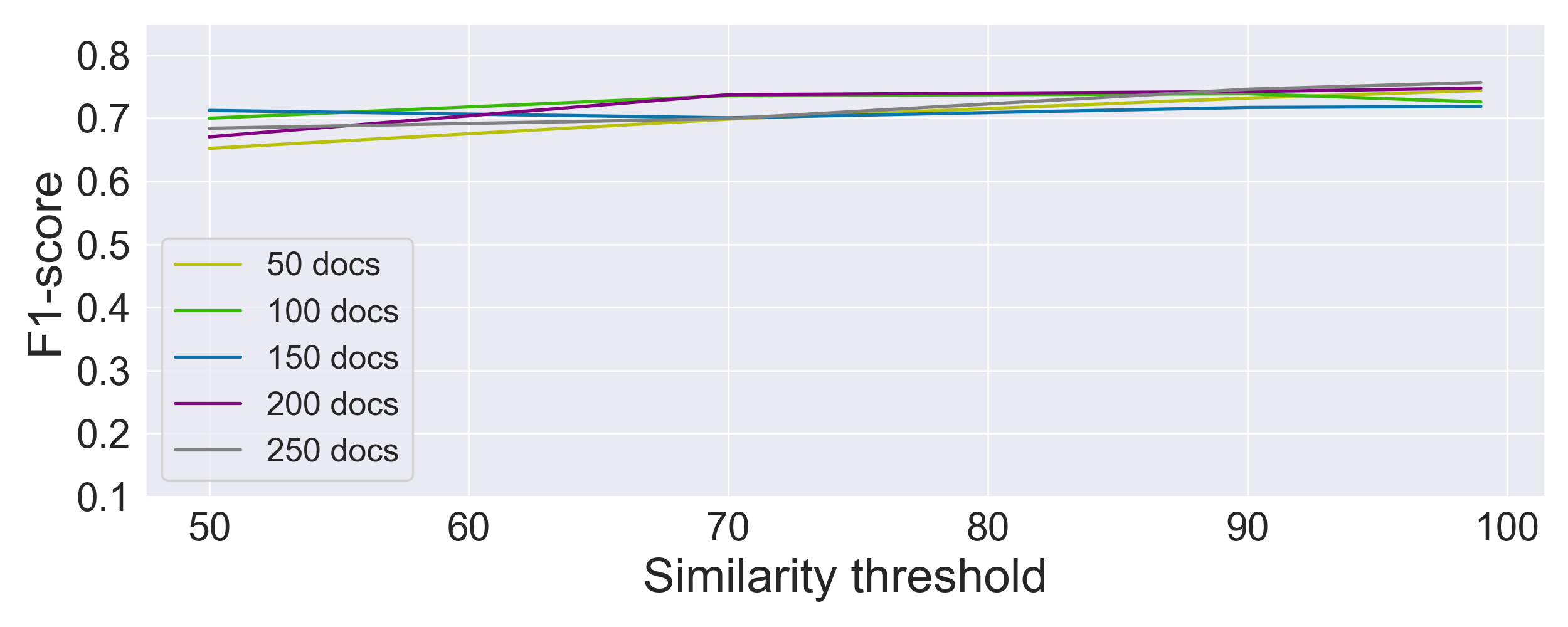}
     \caption{Results on the Twitter dataset using the BERTweet attribution model.}
     \label{fig:gen_test_twitter_BERT_varying_sim}
\end{subfigure}
\begin{subfigure}[b]{0.42\textwidth}
     \centering
     \includegraphics[width=\textwidth]{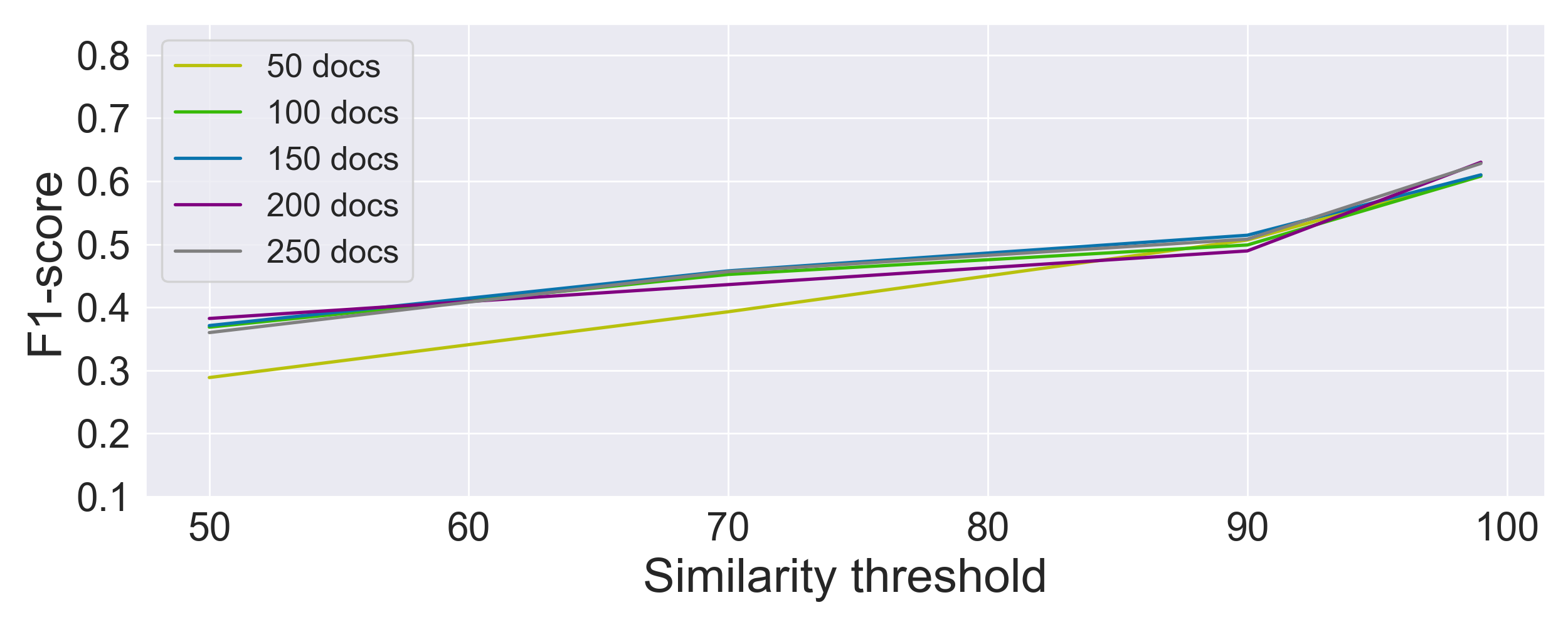}
     \caption{Results on the Twitter dataset using the random forest attribution model.}
     \label{fig:gen_test_twitter_rf_varying_sim}
\end{subfigure}
\caption{Authorship attribution performances for models trained on human-created texts and tested on AI-generated texts, with sub-samples of AI-generated texts that bear more or less similarity to their fine-tuning data (from the target author). Legend indicates different number of fine-tuning documents per author used to generate texts.}
\label{fig:gen_test_varying_sim}
\end{figure*}

Moreover, both RF and BERT record distinctly lower scores for attributing AI-generated texts on the Twitter dataset. However, as the BERT models (particularly BERTweet) still attribute the AI-generated texts to a reasonable degree, this suggests that the AI-generated Twitter texts are still capturing authorship in some manner. Even so, the generated tweets do appear less able to mimic authorship when compared to the generated blog posts.

We also examined the effects of varying the number of AA training documents. We present these results in Fig.~\ref{fig:gen_test_varying_docs}. From this, we note clear similarities to the scores achieved in Fig.~\ref{fig:human_cross_val}. For our RF model, we again find a reasonable degree of stability in its scores on the AI-generated texts. We do, however, note lower performances on the Twitter dataset for AI-generated texts for varying numbers of training documents, indicating that the lower scores in Fig.~\ref{fig:human_train_gen_test_Twitter} are consistent for different numbers of training documents.

We also find that the BERT and BERTweet models again shows less stability, and that it takes more training documents to achieve the higher scores for AI-generated texts when compared to its performances on human-created texts. This points to some degree of separation in the BERT attribution model's performance on AI-generated data versus human-created data. Whilst it may be the model type that most readily assigns AI-generated texts to their target authors, this behaviour only becomes apparent as the numbers of training documents increase. There thus appears to be a `sweet spot' between 150 and 200 documents for both datasets in which the BERT models' performances on human-created data outperform other models, whilst at the same time remaining relatively insensitive to the AI-generated data.

In answer to RQ1, AA models appear to be susceptible to deceptive AI-generated texts, even when relatively small numbers of fine-tuning documents are used. However, we also observe that certain models may prove more resistant to this form of deception. Additionally, we find that this capacity for deception is less evident in our tweet dataset, indicating that this more difficult medium has substantial inhibitory effects on the deceptive abilities of our text generators.

\subsection{RQ2: The Effect of Originality on Authorship Mimicry}

In answer to RQ2, we investigated the degree to which more or less creative AI-generated texts can still deceive our AA models. As the output of the GPT-2 text generators is probabilistic, there is a large degree of scope for the AI-generated data to be highly similar to its source texts or distinctly different. We can thus hypothesise that generated texts that bear a marked similarity to an author's original (human-created) texts have a higher likelihood of being attributed to that author by an AA model. It is thus of interest to account for this and identify whether more `creative' outputs still retain a clear sense of authorship.

We present our results in Fig.~\ref{fig:gen_test_varying_sim}, showing the performance for the BERT (BERTweet for our Twitter dataset, as it was the better performing language model) and RF attribution models on AI-generated texts filtered by a given Levenshtein distance threshold (where the greater the value, the higher degree of similarity is permitted). These results are presented, as with our other experiments, using sets of 5 authors with 300 documents per author, repeated over 20 trials.

From this figure, we can see that on our blog dataset, the similarity of the AI-generated documents to their training data appears to have little tangible impact on its scores. Instead, we find that in general the models still attribute AI-generated texts correctly, even when faced with more original generated texts. This indicates that authorship is retained even when the AI-generated blog texts are reasonably distinct from the documents used to train the generator.

For our Twitter dataset, we see more of an impact on the deceptiveness of our AI-generated texts, with a small decrease in F1-score for the BERTweet model and a fairly substantial decrease in F1-score for the RF model. This further indicates that the GPT-2 generator is less effective at authorial text generations on Twitter data, struggling to capture authorial style when producing more creative tweets.

\begin{figure*}[tb]
\centering
\begin{subfigure}[b]{0.42\textwidth}
     \centering
     \includegraphics[width=\textwidth]{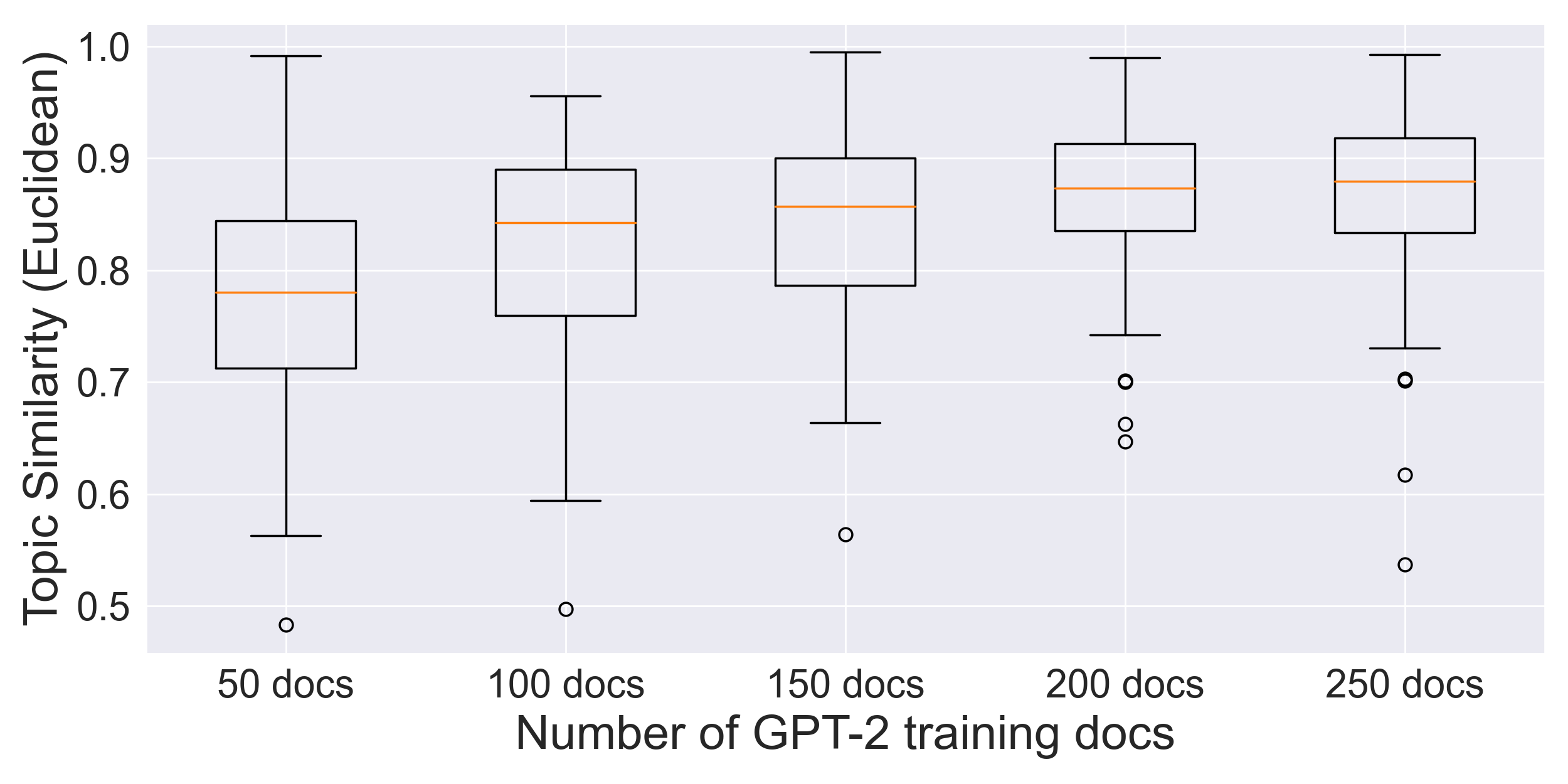}
     \caption{Results for the blog dataset.}
     \label{fig:blog_topics_euclidean}
\end{subfigure}
\quad
\begin{subfigure}[b]{0.42\textwidth}
     \centering
     \includegraphics[width=\textwidth]{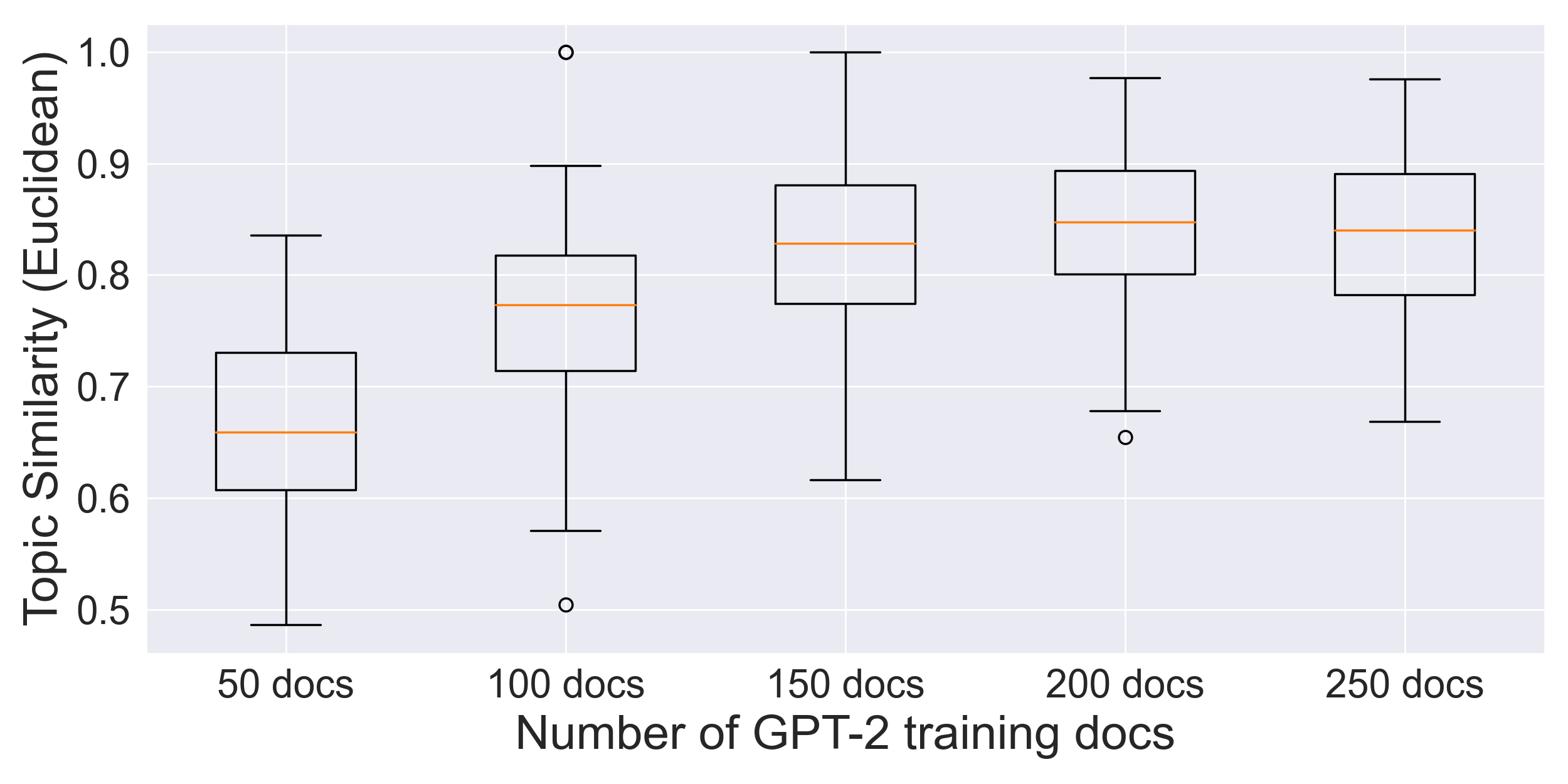}
     \caption{Results for the Twitter dataset.}
     \label{fig:twitter_topics_euclidean}
\end{subfigure}
\caption{Normalised Euclidean similarities of topic probability distributions in human-created and AI-generated posts for a given author.}
\label{fig:topic_analysis}
\end{figure*}

Thus, in answer to RQ2, we find that more creative AI-generated blog texts are still capable of capturing authorial style sufficiently enough for attribution. However, we also note limitations in this capability on our Twitter dataset, indicating that the GPT-2 generator is less capable of mimicking authorship when presented with shorter and likely noisier fine-tuning data.

Beyond this analysis, we were also interested in examining how the overlap in human-created data between the training data for the AA classifier and the fine-tuning data for the GPT-2 generator effected AA deception. To do this, we repeated the experiments presented in Fig.~\ref{fig:human_train_gen_test}, but instead of training each AA classifier using a set of 250 texts that had also been used to fine-tune the GPT-2 model (plus the additional 50 texts) we sampled from each dataset a new set of unseen texts per author that were not used in the fine-tuning process to act as ``unseen'' training data. We thus sampled a new set of 300 posts per user from the blog dataset and 250 tweets per user from the Twitter dataset (we use 250 tweets rather than 300 as the Twitter dataset was limited to 500 tweets per user). This presents a far tougher challenge, as the GPT-2 model is being tested on its ability to mimic authorial style in a more generalised sense in order to successfully deceive a classifier trained on a distinct set of human-created texts. The results of these experiments are presented in Table~\ref{table:unseen_AA_deception_performances}. In the interest of space, we take the mean of each classifier's scores on generated texts produced using different numbers of fine-tuning documents.  

\begin{table}[!htb]
\small
\centering
\begin{tabular}{ c c c c c }
\toprule
\textbf{Dataset} & \textbf{Model} & \textbf{Precision} & \textbf{Recall} & \textbf{F1-Score} \\
\midrule
\multirow{2}{*}{Blog} & BERT (base) & \textbf{0.79} & \textbf{0.79} & \textbf{0.77}\\
& Random forest & 0.72 & 0.7 & 0.7 \\ 
\midrule
\multirow{3}{*}{Twitter} & BERTweet & \textbf{0.72} & \textbf{0.71} & \textbf{0.71}\\
& BERT (base) & 0.63 & 0.62 & 0.62 \\ 
& Random forest & 0.54 & 0.53 & 0.53 \\ 
\bottomrule
\end{tabular}
\caption{Performances for AA models trained on genuine texts and tested on generated texts using unseen human-created training data.}
\label{table:unseen_AA_deception_performances}
\end{table}

From these results, we note a drop in performance across both datasets, indicating that GPT-2's deceptive capabilities will likely be hindered when an unseen training set is used for an AA classifier. This is particularly notable on the Twitter dataset, in which the BERT and RF models typically struggle to attribute the deceptive texts as intended. Whilst we were only able to use 250 training documents on the Twitter dataset, given the significant drop in performance relative to our previous experiments we suspect that it is unlikely that a further 50 tweets (to match our blog experiment) would have made a substantial difference.

This loss in performance is less pronounced on the blog dataset, with the BERT model in particular still showing a good degree of susceptibility to attributing these deceptive texts as intended. It thus appears that GPT-2 is still reasonably able to mimic authorship in this setting, indicating that on the longer-form blog data its has some capabilities of learning the author's style in a more generalised sense, beyond the fine-tuning data. This appears to be less true for the Twitter dataset, though the BERTweet results indicate that some degree of generalisable authorial mimicry is still achieved.

These results are of interest, as they highlight a key weakness in the capabilities of GPT-2 as a tool for authorial deception. Whilst GPT-2 appears able to deceive authorship in cases where there is overlap between its fine-tuning set and an AA classifier's training set, it appears less capable of capturing a generalised sense of authorship sufficient to fool AA classifiers trained on a hidden set of data. This points to a potential solution to safeguarding these AA systems against generative deception, indicating that the developers of these systems could cultivate private sets of data for each relevant author as a means of shielding against these attacks. It is worth noting, however, that in many situations where the AA classifier is solely reliant on public data, such as on Twitter, this creation of a hidden training set may prove difficult. Moreover, the fact that in some settings GPT-2 can still deceive AA models trained on hidden datasets means that further mitigation strategies will likely be needed as the capacity of these generative models continues to grow.

\subsection{RQ3: Comparative Linguistic Analyses}

Finally, in answer to RQ3 we conducted a series of comparative analyses between the human-created and AI-generated texts for each of our authors.

\begin{figure*}[tb]
\centering
\begin{subfigure}[b]{0.42\textwidth}
    \centering
\includegraphics[width=\textwidth]{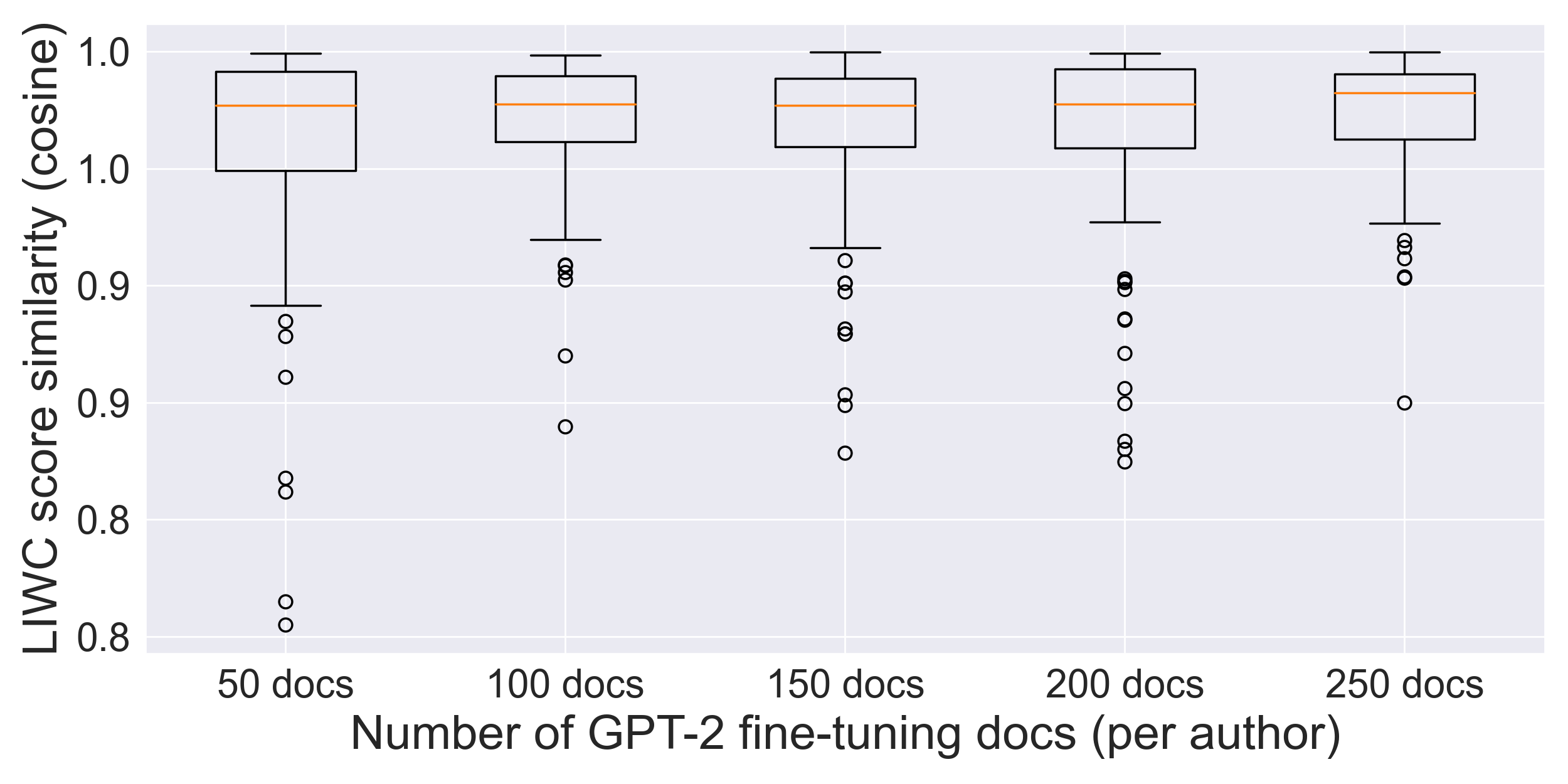}
    \caption{Summary Variables on the blog dataset.}
    \label{fig:summary_variables_blog}
\end{subfigure}
\quad
\begin{subfigure}[b]{0.42\textwidth}
     \centering
     \includegraphics[width=\textwidth]{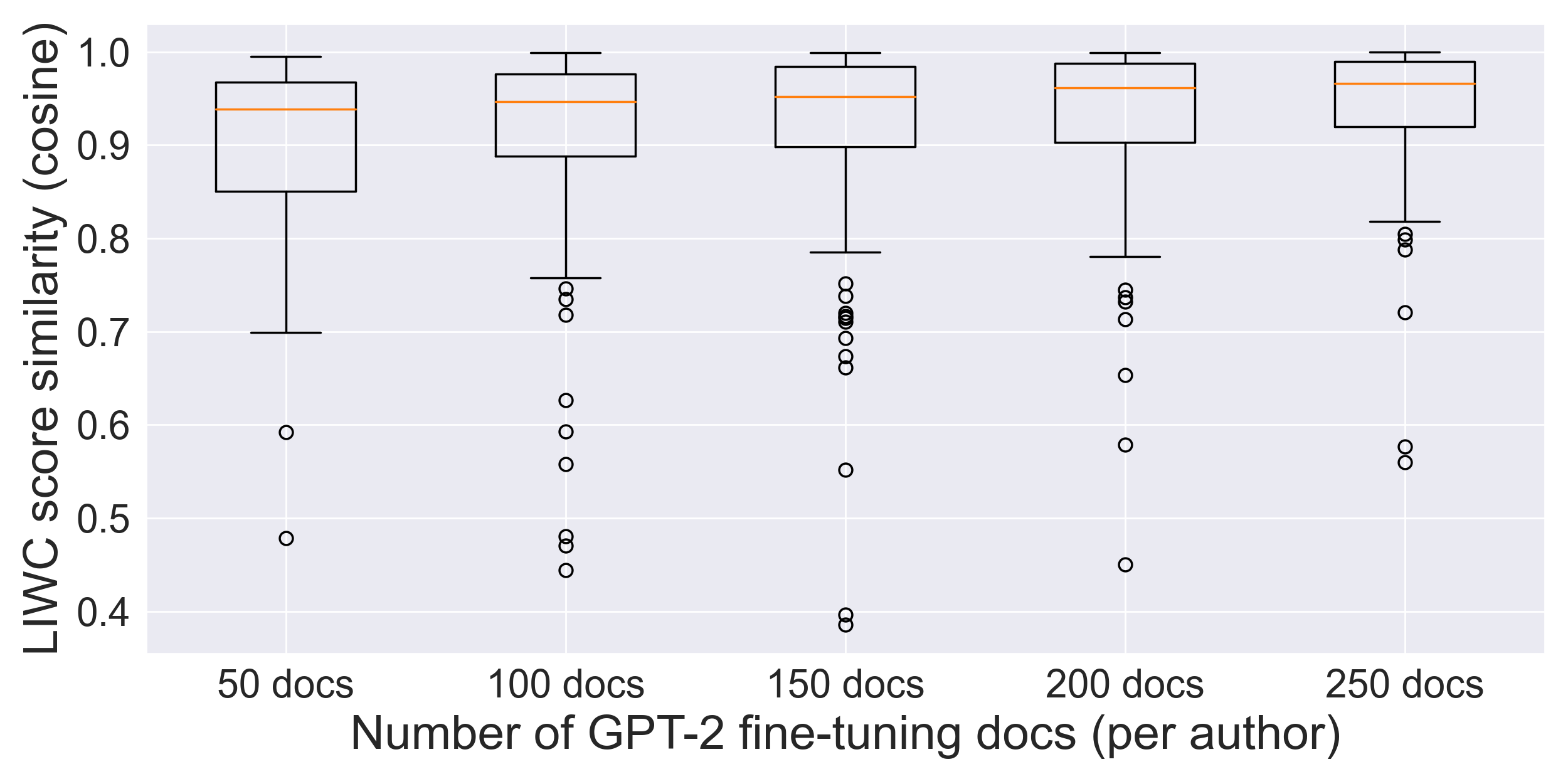}
     \caption{Summary Variables on the Twitter dataset.}
     \label{fig:sumary_variables_twitter}
\end{subfigure}
\begin{subfigure}[b]{0.42\textwidth}
     \centering
     \includegraphics[width=\textwidth]{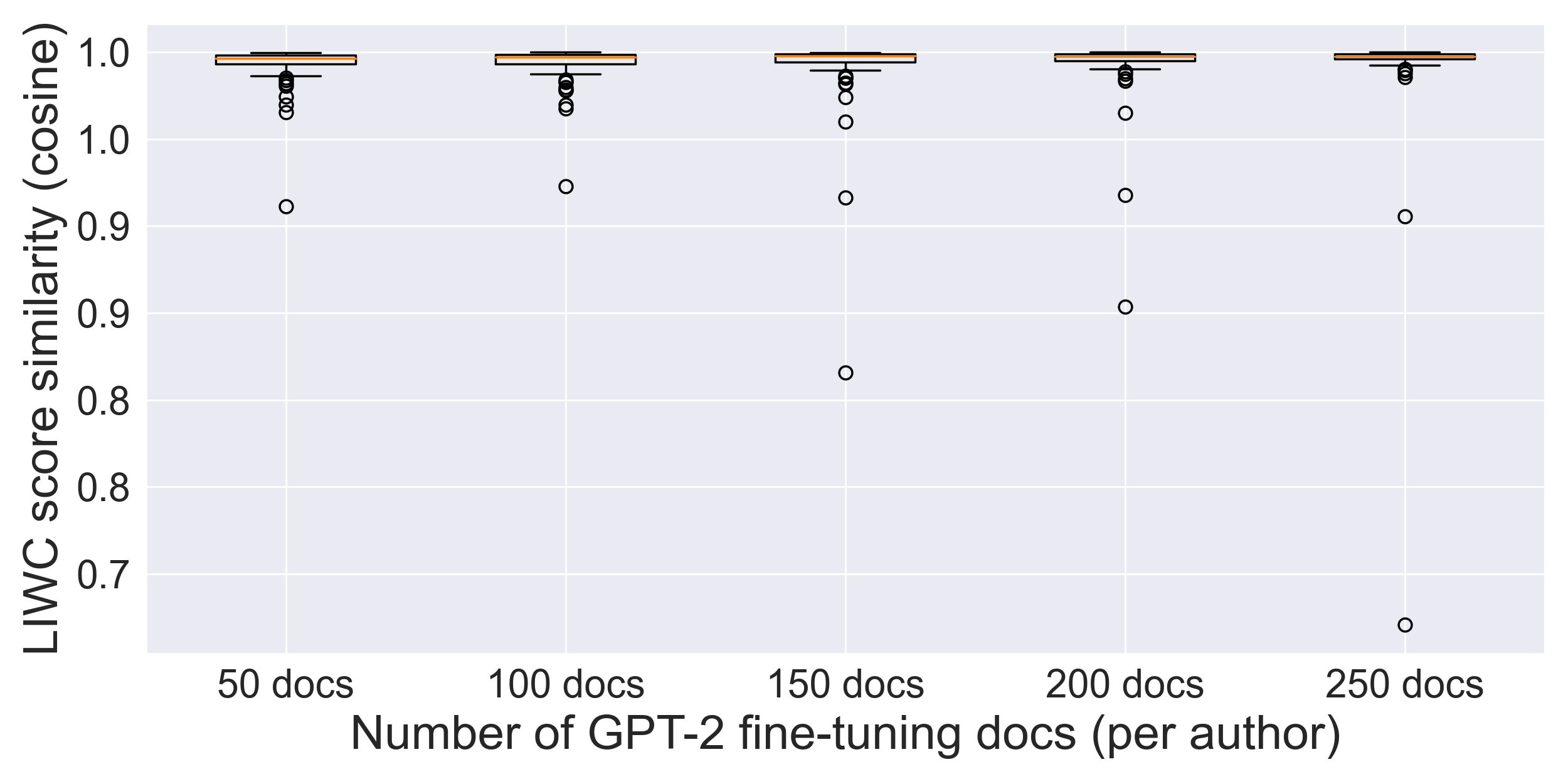}
     \caption{Grammar Other on the blog dataset.}
     \label{fig:language_metrics_blog}
\end{subfigure}
\begin{subfigure}[b]{0.42\textwidth}
     \centering
     \includegraphics[width=\textwidth]{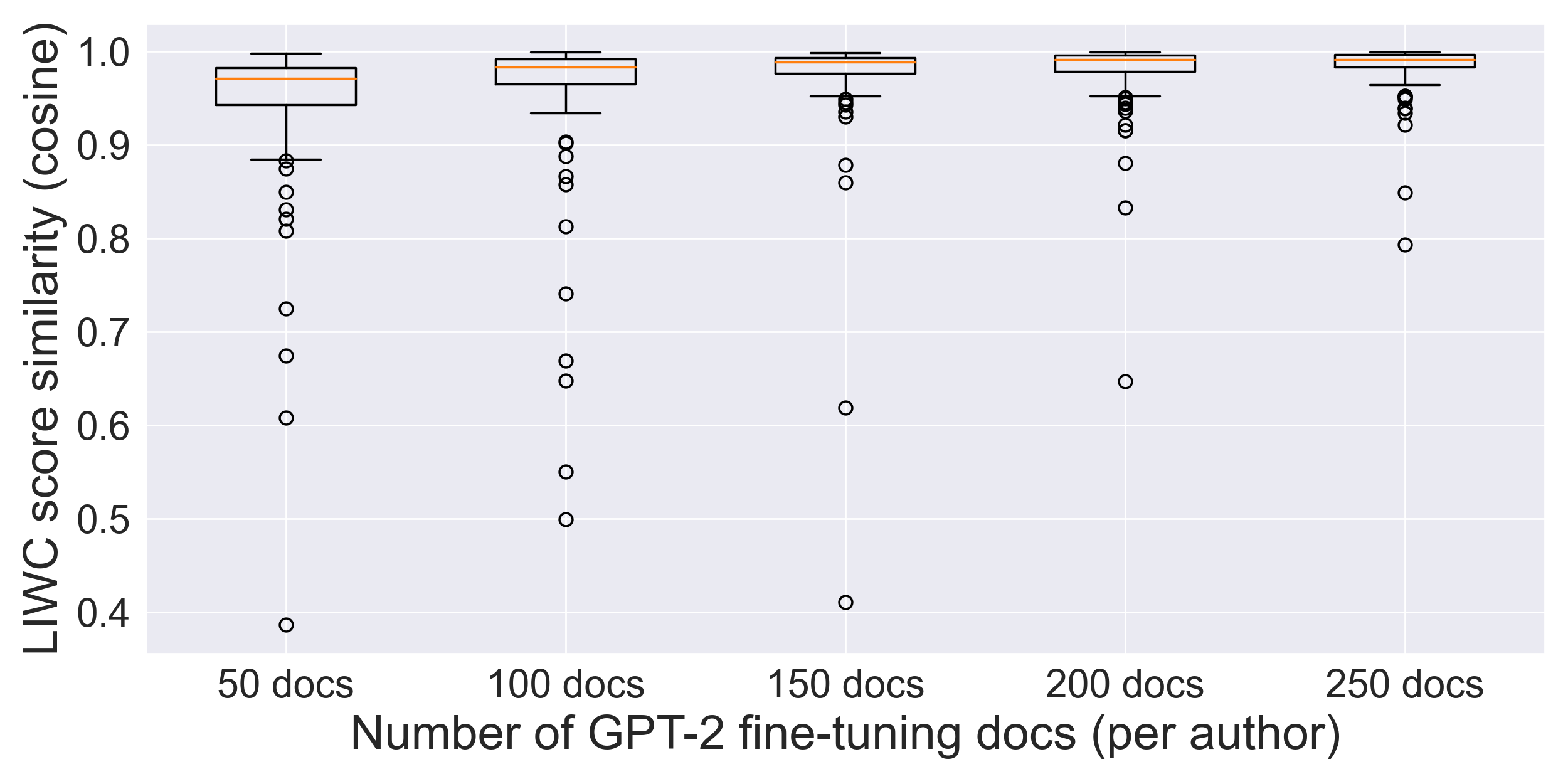}
     \caption{Grammar Other on the Twitter dataset.}
     \label{fig:language_metrics_twitter}
\end{subfigure}
\caption{Cosine similarities in LIWC dimensions scores for AI-generated and human-created posts from the same author.}
\label{fig:liwc_analysis}
\end{figure*}

\subsubsection{Topic Analysis}

In Fig.~\ref{fig:topic_analysis} we present the results of our comparative topic analysis for our two datasets. We record pairwise normalised Euclidean similarities between the 100 authors (5 authors $\times$ 20 trials) tested in our AA experiments, for texts generated with varying numbers of fine-tuning documents. A score of 1.0 indicates maximal similarity and a score of 0.0 indicates no similarity.

From these figures, we find that there is a considerable degree of similarity in the topics captured in AI-generated texts for the blog datasets. We also note a clear upward trend in the degree of topic overlap as the number of fine-tuning documents used to generate our texts is increased. To verify this, we utilise the Wilcoxon signed rank test. This identified statistically significant differences between 50 and 100 fine-tuning documents ($W=1385.0$, $p<0.001$) and between 100 and 200 documents ($W=1521.0$, $p<0.0016$) with a significance level of 0.05.

Similar patterns are noted in the Twitter data, albeit with consistently lower similarities compared to the blog data (potentially an artifact of the shorter, noisier data). Despite this, the AI-generated texts do still show reasonable degrees of similarity in topics when compared to their target author's. We also note similar upward trends in similarity as the number of fine-tuning documents increase. We again utilise the Wilcoxon signed rank test, which noted statistically significant differences between all distributions bar 150 and 200 documents ($W=270.0$, $p=0.094$), 150 and 250 documents ($W=330.0$, $p<0.402$), and 200 and 250 documents ($W=358.0$, $p=0.655$) for our analysis with a significance level of 0.05. 

Interestingly, it is worth noting that whilst clear improvements in the capturing of topics are noted in both datasets as the fine-tuning documents increase, we see less clear evidence in our attribution experiments above (i.e., in Fig.~\ref{fig:human_train_gen_test}) of increases in performance as the number of fine-tuning documents increase. This suggests that semantic overlap between the AI-generated and human-created texts may not be the sole driving factor to successful attribution. 

Based on these findings, we examined the degree to which our AI-generated texts are mimicking the target author's topics of discussion, rather than their innate writing style. To do this, we repeated the experiments discussed in the above sections using only stylometric features to train our AA classifiers. As stylometric features are centred around measuring an author's writing style through their linguistic patterns, rather than the semantic content of their texts, we can reason that if the AI-generated texts were only able to capture the target author's topics an AA classifier trained only on style-based features would not be fooled. We excluded n-gram features in these tests as these are likely to encode both stylistic and semantic aspects of a given text. We were also unable to repeat these experiments using the BERT AA classifier as the embeddings learned during the fine-tuning phase would likely encode topics as well as style.

Through these experiments, we find little to no change in the performances of the AA classifiers (and thus their susceptibility to AI-generated deceptive texts) on either dataset, both when human-created texts were used for both training and testing and when AI-generated deceptive texts were used in testing. This indicates that, whilst the AI-generated texts are able to consistently encode the topics of interest present in the fine-tuning data, they also capture authorial style to the degree that mimicry can still be achieved when AA models are trained to classify based on stylistic features only.

\subsubsection{LIWC Analysis}

In Fig.~\ref{fig:liwc_analysis}, we report the cosine similarity between LIWC dimensions scores for the ``Grammar Other'' and ``Summary Variables'' dimensions for both of our datasets. We report these dimensions as they offer the most relevant comparisons of both the syntactical and psycho-linguistic similarities between our AI-generated and human-created texts. 

The ``Grammar Other'' dimension provides measures of grammar and syntax-based proprieties of a text, including measures verb usage, adjectives, and interrogatives, whilst the ``Summary Variables'' dimension provides measures of the degree of analytical thinking, emotional tone, and authenticity~\cite{tausczik2010}. Whilst we report results on these two dimensions, similar patterns of similarity between AI-generated and human-created texts were recorded for all LIWC dimensions.

From this, we note high degrees of similarity for both LIWC dimensions across both datasets, regardless of the number of fine-tuning documents used. It appears that the GPT-2 text generators show a strong ability towards capturing the lower level, latent aspects of language usage present in the fine-tuning texts of the target authors. Additionally, they prove capable of doing this even with relatively few fine-tuning documents.

The consistent manner in which the text generators can capture these authorial style patterns goes some way to explaining why the tested AA models are consistently duped by these AI-generated texts. Even the classical machine learning models are largely leveraging low-level text-based features. Whilst these features may allow for good AA performances, they focus on aspects of authorship that AI-based text generators are particularly suited to mimicking, thereby enhancing their vulnerability.

\subsection{Limitations}

There are some limitations to this study that bear mentioning. At this stage, our work is focused on the viability of natural language models in mimicking a target author's writing style and thus achieving fair-to-good performances is sufficient. Further research into optimising the model performance and examining the effects this has on AA of deceptive AI-generated texts would be of interest. Experimenting with more powerful natural language models (e.g., GPT-3) would also be valuable.

Additionally, the results recorded here should be considered an initial examination of the viability of AI-based AA deception. Further, large-scale studies that examine a larger number of authors across multiple large datasets using different text generators will be of great use in further validating our findings and providing a more generalised understanding of this threat. It is worth noting that given the computational expense required to fine-tune transformers en masse this is not a trivial undertaking.

It is also worth noting that whilst LDA and LIWC provide powerful capabilities to analyse text at scale, granular accuracy can be difficult to achieve. Therefore, a complementary qualitative-based analysis of AI-generated texts as they compare to human-created texts would be of great value. Additionally, although we utilise LDA due to its established position in the online research community, experiments with more state-of-the-art topic modelling approaches (e.g., BERTopic) may warrant further investigation.

\section{Broader Perspectives \& Ethics}

Through our experiments, we find that powerful generative models are capable of mimicking authorial style sufficiently to capture both their linguistic patterns and their topics of discussion. Additionally, we find that this mimicry is sufficiently convincing that AI-generated deceptive texts are capable of deceiving trained AA classifiers. This seems particularly true for longer texts, such as blog-style data, where successful deception can often be achieved with minimal fine-tuning. Given this capacity for deception, it is thus possible that these capabilities could be leveraged to do harm.

However, it is important to note that currently it is still possible to distinguish between AI-generated texts and human-crafted texts when one is actively looking for them, with classifiers trained to this task achieving fair-to-good performances~\cite{jawahar2020}. Given this, it is likely that AA systems in current use can mitigate the threats of NLG-based deception by being combined with some form of classifier trained specifically to the task of detecting AI-generated content. The downside is that these forms of AI-generated text detectors are often limited in their generalisability and are typically only able to perform well on the platform (e.g., Twitter, blog) and model-type (e.g., GPT-2, XLM) they are trained on~\cite{jawahar2020}. This may mean that, in many cases, a bespoke detector model would need building, alongside the AA classifier, thus increasing the barrier for developing trustworthy AA systems. Ultimately, given the capabilities of authorship deception demonstrated here, it is important that designers of AA systems take into consideration the potential that these systems could be attacked by deceptive NLG texts and reflect on whether additional NLG detection strategies are needed.

\subsection{Ethical Considerations}

We made sure to take suitable steps in our data collection and analysis to ensure an ethical study and preserve user privacy. All data extractions were made in accordance with Twitter's Standard API terms and conditions~\cite{TwitterAPI}, with no deleted, protected, or suspended accounts included in our analysis. Additionally, in order to avoid issues of account or user identification we do not name any accounts in this paper, relying solely on aggregate data analysis to protect user privacy for both datasets used (i.e., blog and Twitter). For similar reasons, we also refrain from using verbatim quotations from user texts within our paper.

\section{Conclusions and Future Work}

In summary, we find that AI-based text generators are capable of generating online texts that can deceive AA models. We also find that these abilities towards authorial deception are maintained amongst the more original outputs of our GPT-2 model for our blog dataset, indicating that the capturing of authorial style is not simply the product of the text generator closely paraphrasing posts provided during fine-tuning. We also note limitations when attempting to mimic tweet authorship indicating that powerful NLG models are still limited by shorter-form and typically noisier social media texts. Further limitations are also observed when hidden training sets are used in the development of AA classifiers, indicating that GPT-2 struggles to produce texts with a generalisable sense of authorship. This finding indicates that the use of hidden training sets may be a viable solution to safeguarding against these deceptive acts, particularly when using short-from Twitter data.

Given that many AA systems have been proposed for critical tasks such as spam detection~\cite{duman2016} and forensic investigation~\cite{perkins2018}, the capacity for powerful text generators to mimic a given author could pose a significant threat to these real-world systems. Whilst current text generators are limited in the degree to which users can control the generated text output, improvements in this space are coming rapidly~\cite{dathathri2019}. These models could then leverage the capabilities towards authorial mimicry noted in this paper and use them to disrupt existing AA systems.

Further research into the capabilities and ethical problems posed by these models is thus needed whilst generated texts are still (relatively) easily identifiable and whilst the capacity towards authorial mimicry is still limited by the model's inconsistent outputs. This will necessitate further research into the capacity of different natural language models to mimic authorship, alongside studies of authorship mimicry on other online platforms. Complementary research examining the degree to which these deceptive texts can fool human evaluators would also be of interest.

Additional research into the degree to which current approaches to steering AI-generated text outputs are capable of retaining authorial style will also be needed. Our findings raises the question of whether style-transfer or topic-controlled generation could be leveraged to generate texts in a given authors style that also contain some form of content (e.g., topic, sentiment) determined by the deceiver. If possible, this would further demonstrate the powers of these NLG models and emphasise their potential for malicious use.

\fontsize{9.0pt}{10.0pt} 
\selectfont
\bibliography{FULL-JonesK-Shortened} 

\end{document}